\documentclass[journal]{IEEEtran}
\IEEEoverridecommandlockouts
\usepackage{cite}

\usepackage{graphicx}
\usepackage{amsmath}
\usepackage{amssymb}
\usepackage{booktabs}
\usepackage{amsfonts}
\usepackage{textcomp}
\usepackage[table]{xcolor}
\usepackage{hyperref}
\usepackage{float} 
\usepackage{subcaption}
\usepackage{algorithm}
\usepackage{algcompatible}
\usepackage[noend]{algpseudocode}
\usepackage{soul}
\usepackage{subfiles}
\usepackage{multirow}
\usepackage{hhline}
\usepackage{xpatch}
\usepackage{balance}
\usepackage{placeins}
\usepackage{tikz}
\usepackage{tabularx}
\usepackage{bm}
\usepackage{boldline}

\def\BibTeX{{\rm B\kern-.05em{\sc i\kern-.025em b}\kern-.08em
    T\kern-.1667em\lower.7ex\hbox{E}\kern-.125emX}}
    
\algnewcommand\algorithmicforeach{\textbf{for each}}
\algdef{S}[FOR]{ForEach}[1]{\algorithmicforeach\ #1\ \algorithmicdo}

\makeatletter
\xpatchcmd{\algorithmic}{\itemsep\z@}{\itemsep=.03cm}{}{}
\makeatother

\begin{document}

\author{
\IEEEauthorblockN{Tharindu Samarakoon,}
\and
\IEEEauthorblockN{Kalana Abeywardena,}
\and
\IEEEauthorblockN{and Chamira U. S. Edussooriya}

}


\title{Arbitrary Volumetric Refocusing of Dense and Sparse Light Fields}

\maketitle


\begin{abstract}
A four-dimensional light field (LF) captures both textural and geometrical information of a scene in contrast to a two-dimensional image that captures only the textural information of a scene. Post-capture refocusing is an exciting application of LFs enabled by the geometric information captured. Previously proposed LF refocusing methods are mostly limited to the refocusing of single planar or volumetric region of a scene corresponding to a depth range and cannot simultaneously generate in-focus and out-of-focus regions having the same depth range. In this paper, we propose an end-to-end pipeline to simultaneously refocus multiple arbitrary planar or volumetric regions of a dense or a sparse LF. We employ pixel-dependent shifts with the typical shift-and-sum method to refocus an LF. The pixel-dependent shifts enables to refocus each pixel of an LF independently. For sparse LFs, the shift-and-sum method introduces ghosting artifacts due to the spatial undersampling. We employ a deep learning model based on U-Net architecture to almost completely eliminate the ghosting artifacts. The experimental results obtained with several LF datasets confirm the effectiveness of the proposed method. In particular, sparse LFs refocused with the proposed method archive structural similarity index higher than $0.9$ despite having only $20\%$ of data compared to dense LFs.


\end{abstract}


\section{Introduction}
A four dimensional (4-D) light field (LF) is a simplified form of the seven-dimensional plenoptic function~\cite{Ade1991,Zhan2003} that completely describes the light emanating from a scene. An LF captures both geometric and textural information of a static scene in contrast to a two-dimensional (2-D) image which captures only textural information of a static scene. These additional information available with an LF paves the way to accomplish novel tasks such as depth estimation~\cite{wanner2013variational,tao2013depth,wang2015depth} and occlusion suppression~\cite{vaish2004occlusion,Dan2007,Liyanageocclusion} that are not generally possible with 2-D images. Furthermore, LFs have been employed for multiple applications in computer vision systems such as mobile robotics~\cite{Dong2013,Tsai2017,Baj2018} and underwater imaging~\cite{Piza2013,Dan2014}.


\begin{figure}[!t]
    \centering
    \begin{subfigure}[c]{0.49\columnwidth}
    \vspace{1mm}
        \centering
         \includegraphics[width=\textwidth]{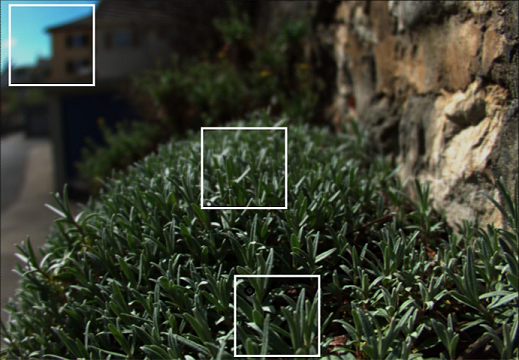}
        \label{fig:single_depth_refocus_img}
    \end{subfigure}
    \begin{subfigure}[c]{0.49\columnwidth}
    \vspace{-3mm}
        \centering
        \includegraphics[width=\textwidth]{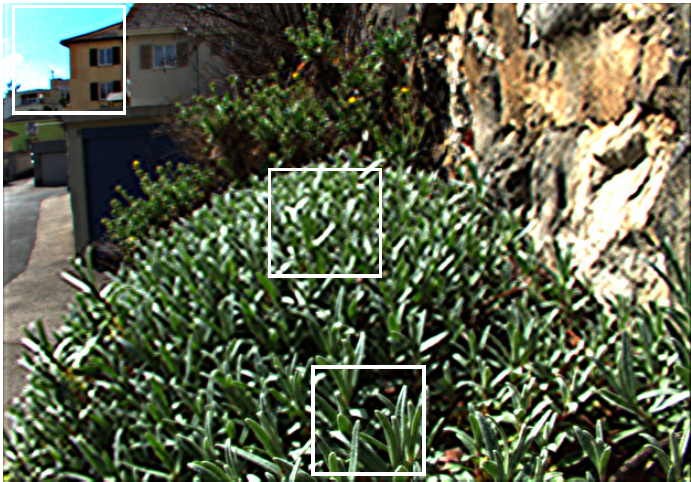}

        \label{fig:multi_depth_sakila_akka}
    \end{subfigure}
    
    \vspace{-3mm}
    \begin{subfigure}[c]{0.49\columnwidth}
        \centering
        \resizebox{1\columnwidth}{!}{%
            \centering
            \includegraphics[height=3cm]{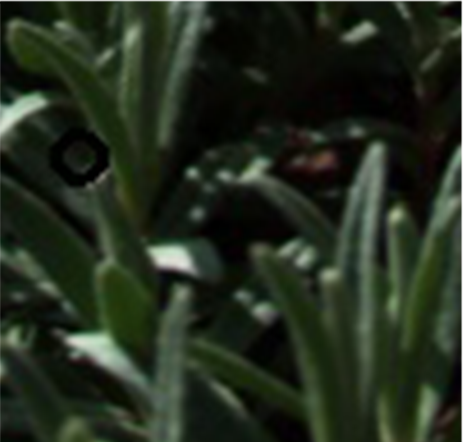}
            \includegraphics[height=3cm]{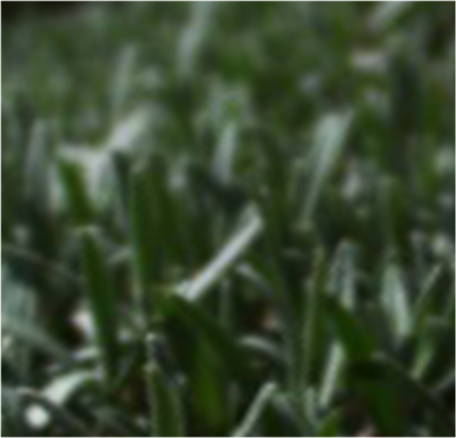}
            \includegraphics[height=3cm]{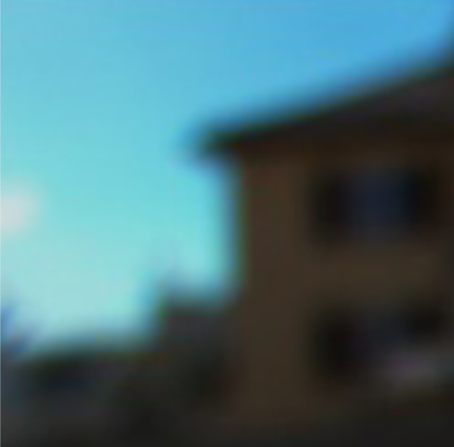}

            \hspace{-3mm}
        }
        \caption{}
        \label{fig:single_depth_refocused_img_slices}
    \end{subfigure}
    \begin{subfigure}[c]{0.49\columnwidth}
        \centering
        \resizebox{1\columnwidth}{!}{%
            \centering
            \includegraphics[height=3cm]{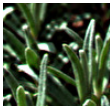}
            \includegraphics[height=3cm]{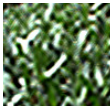}
            \includegraphics[height=3cm]{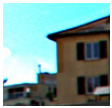}

            \hspace{-3mm}
        }
        \caption{}
        \label{fig:multi_depth_sakila_akka_refocused_img_slices}
    \end{subfigure}
    
    \begin{subfigure}[c]{0.49\columnwidth}
    \vspace{1mm}
        \centering
         \includegraphics[width=\textwidth]{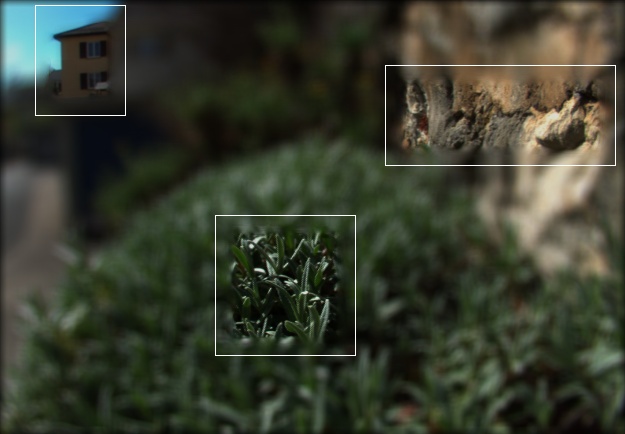}
        \label{fig:dense_multi_refocused_img}
    \end{subfigure}
    \begin{subfigure}[c]{0.49\columnwidth}
    \vspace{-3mm}
        \centering
        \includegraphics[width=\textwidth]{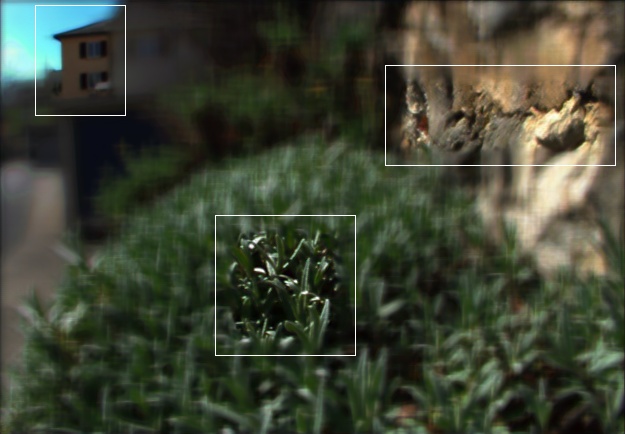}

        \label{fig:sparse_multi_refocused_img}
    \end{subfigure}
    
    \vspace{-3mm}
    \begin{subfigure}[c]{0.49\columnwidth}
        \centering
        \resizebox{1\columnwidth}{!}{%
            \centering
            \includegraphics[height=3cm]{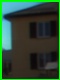}
            \includegraphics[height=3cm]{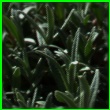} 
            \includegraphics[height=3cm]{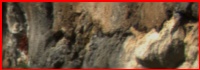}
            \hspace{-3mm}
        }
        \caption{}
        \label{fig:dense_multi_refocused_img_slices}
    \end{subfigure}
    \begin{subfigure}[c]{0.49\columnwidth}
        \centering
        \resizebox{1\columnwidth}{!}{%
            \centering
            \includegraphics[height=3cm]{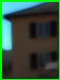}
            \includegraphics[height=3cm]{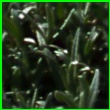} 
            \includegraphics[height=3cm]{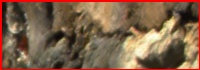}
            \hspace{-3mm}
        }
        \caption{}
        \label{fig:sparse_multi_refocused_img_slices}
    \end{subfigure}

    \caption{Refocusing of ``Bush" LF; (a) single planar refocus~\cite{ng2005light}; (b) two volumetric regions~\cite{multivolume}; (c) proposed arbitrary volumetric refocusing using a dense LF; (d) proposed arbitrary volumetric refocusing using a sparse LF having cross-shaped SAIs. For (c) and (d), narrow-depth and wide-depth regions are shown in boxes with green- and red-colored outlines, respectively.}
    \label{fig:dense_sparse_multi_refocus}
\end{figure}

\begin{figure}[t]
    \centering
    \resizebox{1\columnwidth}{!}{%
        \centering
        \subcaptionbox{\label{fig:camera_array}}{\centering \includegraphics[height=3cm]{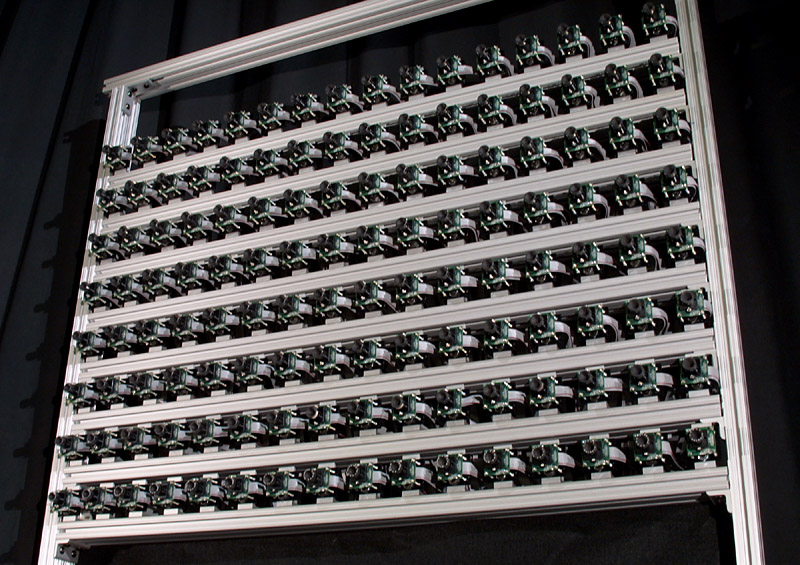}}
        \subcaptionbox{\label{fig:lytro_camera}}{\centering \includegraphics[height=3cm]{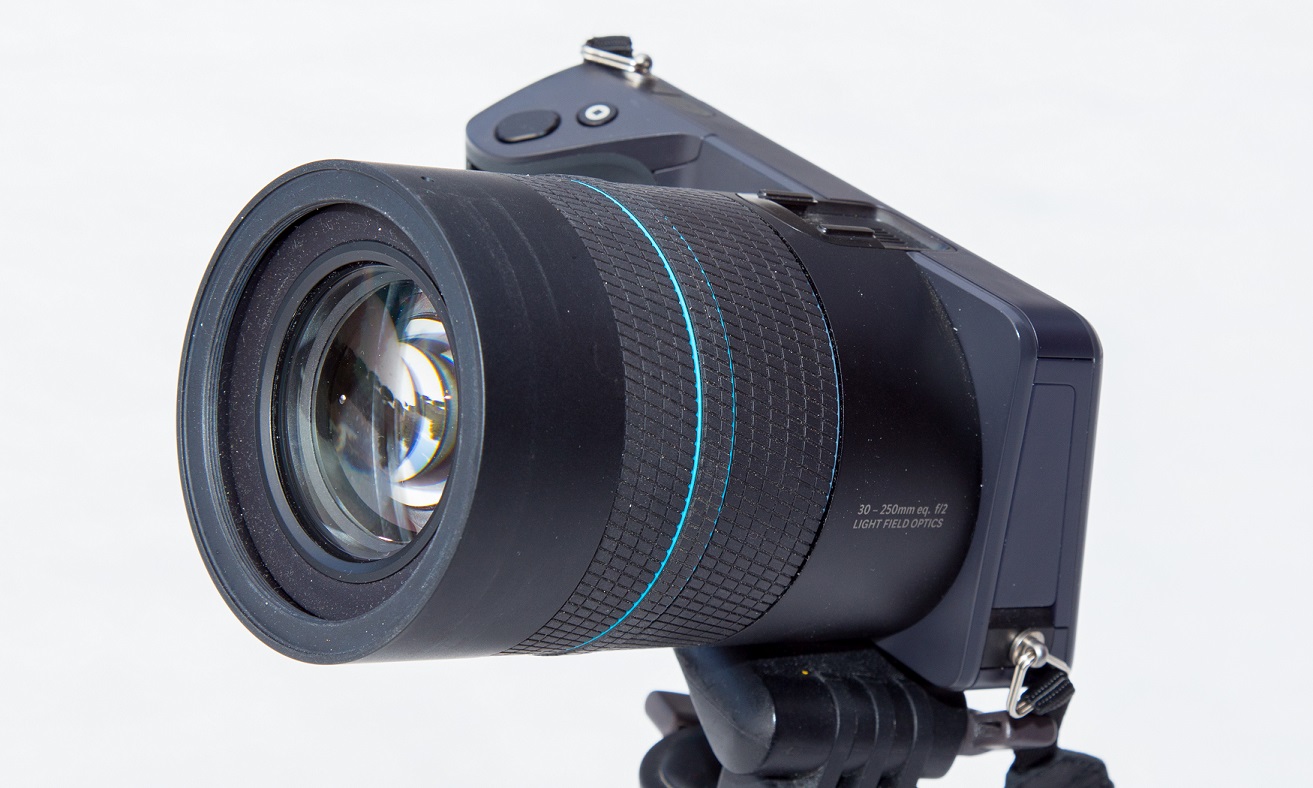}}
        \subcaptionbox{\label{fig:epiModule}}{\centering \includegraphics[height=3cm]{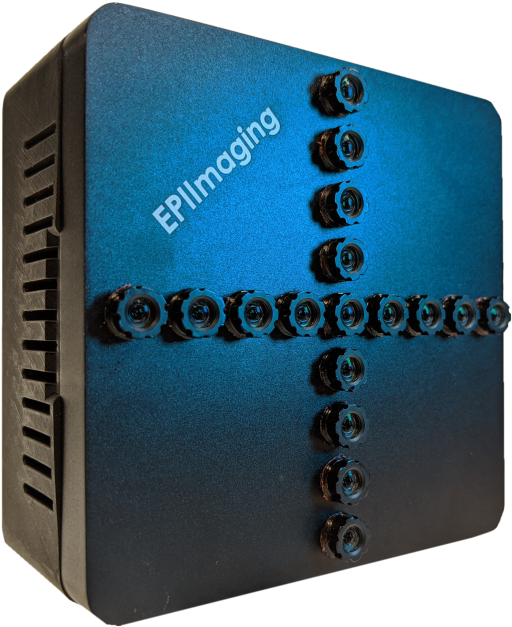}}
    }
     \hfill

    \caption{Different types of LF cameras; (a) LF video camera array~\cite{LFVideoCamera}, (b) Lytro Illum dense LF camera from Lytro, Inc, (c) EPIModule Sparse LF camera from EPIImaging LLC.}
    \label{fig:LFcameras}
\end{figure}

Post-capture refocusing---ability to change the focused regions of a scene after capturing---is another exciting application of LFs. This was first demonstrated by Ng \textit{et al.} in the seminal work~\cite{ng2005light}, where refocusing over a narrow depth range (called planar refocusing) was achieved by appropriately shifting the pixels of sub-aperture images (SAIs) of an LF and subsequently summing these shifted SAIs. The required shifts are determined by the depth (or disparity) of the region to be focused. Even though planar refocusing created a paradigm shift in LF photography, inability to refocus over a wide depth range was a drawback. Refocusing over a wide depth range (called volumetric refocusing) was demonstrated by Dansereau \textit{et al.} in~\cite{volumetric} using a 4-D linear and shift-invariant filter having a hyperfan passband. The depth range of a scene that should be refocused is determined by the angle of the hyperfan, and the planar refocusing is a special case of the volumetric refocusing. Jayaweera \textit{et al.}~\cite{multivolume} demonstrated multi-volumetric refocusing, where multiple regions having wide depth ranges were simultaneously refocused using a 4-D linear and shift-invariant filter having multiple hyperfan passbands. Such multi-volumetric refocusing adds novel features in LF photography and cinematography~\cite{Tro2019}. Furthermore, multi-volumetric refocusing can achieve single planar or volumetric region as a special case. However, all these methods cannot refocus a planar or volumetric region while keeping a planar or volumetric region having same depth range out-of-focus. In other words, these methods refocuses whole region corresponding to a given depth range, as shown in Fig.~\ref{fig:single_depth_refocused_img_slices} and \ref{fig:multi_depth_sakila_akka_refocused_img_slices}. Furthermore, all these methods have been developed for dense light fields, captured from camera arrays or lenselet-based LF cameras (shown in Figs.~\ref{fig:camera_array} and \ref{fig:lytro_camera}, respectively) leading to higher memory and computational complexities.

Sparse LFs consisting of a subset of SAIs of dense LFs drastically reduces the amount of data captured. For example, a cross-shaped sparse LF captured with a sparse camera array EPIModule shown in \figurename~\ref{fig:epiModule} consists of only $17$ SAIs whereas its dense counterpart consists of $81$ SAIs leading to $79\%$ reduction in data. Such sparse LFs pave the way to implement LF processing methods in resource-constrained devices to achieve real-time processing with low energy consumption. Sparse LFs have been employed in several computer vision applications~\cite{jiang2018sparse_depth,learning_view_synthesis}. Furthermore, several methods to refocus sparse LFs~\cite{2015fast_real,alain2021spatio_sparse} have recently been developed. However, all these methods achieve only the planar refocusing for a whole narrow depth range. Therefore,
novel methods for refocusing of multi volumetric regions of a sparse LF without generating dense LFs as intermediate step is essential to fully exploit the advantages of sparse LFs in LF photography and cinematography.    

In this paper, we propose an end-to-end pipeline to simultaneously refocus multiple arbitrary planar or volumetric regions of both dense and sparse LFs. Our method employs the shift-and-sum method proposed in \cite{ng2005light} for dense LFs. However, we employ \emph{pixel-dependent} depth (or disparity) to determine the shifts of pixels of SAIs in contrast to the pixel-independent depth employed in \cite{ng2005light} to determine shifts. This enables to achieve pixel-dependent refocusing, where a desired planar or volumetric region can be refocused while keeping another region having the same depth range out-of-focus, as shown in Figs.~\ref{fig:dense_multi_refocused_img_slices} and \ref{fig:sparse_multi_refocused_img_slices}. To the best of our knowledge, our method is the \emph{first refocusing method} that achieves simultaneous refocusing of the interested regions while keeping other regions having the same depth ranges out-of-focus in LFs. We then extend the proposed refocusing method to \emph{sparse LFs} having \emph{cross-shaped} SAIs, similar to an LF captured from EPIModule shown in \figurename~\ref{fig:epiModule}. Due to the sparsity of the LF, the shift-and-sum method produces ghosting artifacts in out-of-focus regions even though in-focus regions appear without noticeable artifacts. We employ a U-Net~\cite{unet} based deep learning model to almost completely eliminate the ghosting artifacts, producing refocused sparse LFs having almost the same fidelity as those obtained with dense LFs. Furthermore, our method works as an end-to-end pipeline. That is, once a user selects one or more regions to refocus, our method first generates the depth map of the LF. The required shifts for dense LFs are generated using the generated depth map. For sparse LFs, our method determines the required shifts through an exhaustive search from a feasible range of depths, i.e., without explicitly employing a depth map. The experimental results obtained with several LF datasets confirm the effectiveness of the proposed method. Furthermore, our method is fast and the processing time for a dense LF of size of the $9\times9\times512\times512$ dataset is $3.64$ s, and that for the corresponding cross-shaped sparse LF is $0.71$ s.

The rest of this paper is organized as follows. In Sec.~\ref{sec:relatedWork}, we review the background on existing LF refocusing algorithms. A brief review of the shift-and-sum LF refocusing algorithm and its simplification for digital domain is presented in Sec.~\ref{sec:shiftSumReview}. Then, we describe the proposed arbitrary multi-volumetric LF refocusing method in Sec.~\ref{sec:methodology}. In Sec.~\ref{sec:experiments}, we present the experimental results. Finally, we conclude and present future work in Sec.~\ref{sec:conclusion}.

\section{Related Work}\label{sec:relatedWork}
 
Refocusing of an LF is first demonstrated by Ng \textit{et al.}~\cite{ng2005light}, where SAIs of an LF are shifted appropriately and summed to generate a refocused LF. Here, refocusing was limited to a narrow depth range parallel to the camera plane. This method was further extended to be implemented in the frequency domain in \cite{ng2005fourier}. They showed that the shift-and-sum in the spatial domain~\cite{ng2005light} is equivalent to the generalized Fourier slice theorem in the frequency domain. Furthermore, with the pre-computed spectra on LFs, the computational complexity can be substantially reduced compared to the shift-and-sum method thanks to the fast Fourier transform. Generalizing shift-and-sum method, Martin \textit{et al.}~\cite{tilt_shift} achieved refocusing along non-parallel planes to the camera plane. Furthermore, their method is interactive and the parameters related to the user selected refocus plane are found using a pre-computed depth information.

Refocusing of wide-depth ranges was first proposed in \cite{volumetric} using a 4-D linear and shift-invariant hyperfan filter. Rather than shifting-and-summing, the 4-D filter was designed so that the passband occupies the spectral region corresponding to the depth range needs to be focused, and the stopband occupies the other depth ranges. The width the focused depth range is determined by the angle of the hyferpan-shaped passband. Reducing computational complexity further, \cite{volumeSparseFIR} introduced a 4-D sparse hyperfan filter for volumetric refocusing. Breaking the limit of refocusing to a single depth range, \cite{multivolume} introduced LF refocusing on multiple volumetric regions, allowing to refocus at several objects at different depth ranges. In all these methods, if the objects that should not be focused are within the same depth range that the LF is refocussed, then they too will be in focus. Our approach  overcomes this issue. Specially, it can refocus LFs at any arbitrary-volume or multiple arbitrary-volumes which opens new avenues in LF photography. 


Sparse LFs become more common due to their advantages over dense LFs.
However, sparse LF refocusing requires more sophisticated algorithms to overcome the aliasing artifact due to undersampling. In \cite{2015fast_real}, SAIs of sparse LFs are logically interpolated to generate the missing SAIs during the refocus. In \cite{alain2021spatio_sparse}, an anti-aliasing filter is added to conventional shift-and-sum LF refocusing algorithm to reduce aliasing. In our sparse LF multi arbitrary-volume refocusing method, much faster image restoration U-Net CNN~\cite{unet} is used to address this problem.

The recent development in deep learning paved ways to significantly improve the LF applications. For example, \cite{2021unsupervised} demonstrated depth estimation and visual odometry, using a sparse LF camera and unsupervised learning. More specifically, \cite{deep_sparse} proposes a CNN-based LF refocusing algorithm using only four shifted SAIs. Using a CNN, \cite{machinelearning_multi_refocus} extract 16 refocused images at different depths at once, from a $7 \times 7$ LF. These methods can perform in real-time with low computational complexity. However, unlike to our method, they are not capable of refocusing LFs at multiple regions simultaneously.

\section{Review of Light Field Refocusing using the Shift-and-Sum Algorithm}
\label{sec:shiftSumReview}
\begin{figure}[!t]
    \centering
    \includegraphics[width=0.48\textwidth]{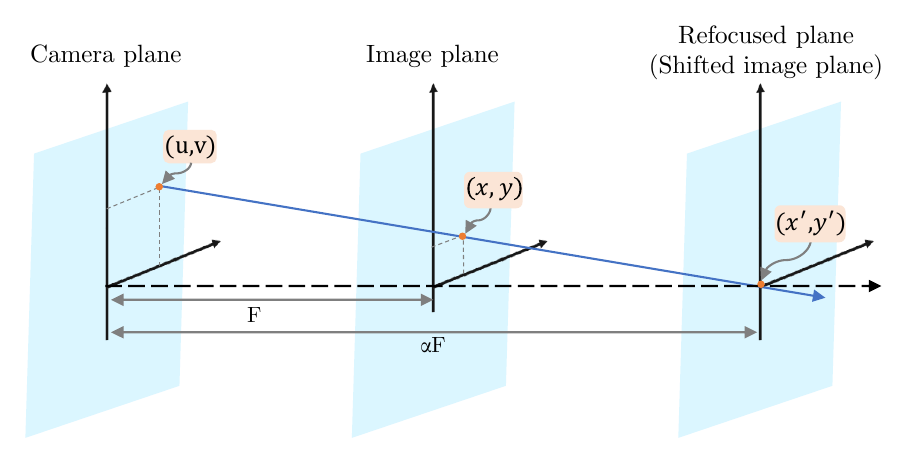}
    \caption{The two-plane parameterization of an LF. The $uv$ plane is called the camera plane, and the $xy$ plane is called the image plane. The distance between the two planes are $F$, and the scene at a distance $F$ is in focus while the rest are out of focus. The image plane is moved artificially to the $x'y'$ plane (called refocused plane) to refocus to a new depth $\alpha F$.}
    \label{fig:refocus}
\end{figure}

We briefly review the shift-and-sum algorithm~\cite{ng2005light} employed for LF refocusing in this section. To this end, we first consider the two-plane parameterization~\cite{levoy1996light} of a continuous-domain LF $L_c(u,v,x,y)$ as shown in Fig~\ref{fig:refocus}, where $(u,v,x,y)\in\mathbb{R}^4$. Here, the $uv$ plane and the $xy$ plane denote the camera plane and the image plane of an LF camera, respectively. The distance between the $uv$ and $xy$ planes is $F$. In generating a focused 2-D image using the captured LF $L_c(u,v,x,y)$, the intensity $\mathcal{I}_c(x,y)$ of the 2-D image at the image plane can be expressed as~\cite{ng2005light}
\begin{align}
\label{eq:LF_intensity}
\mathcal{I}_c(x,y) =\frac{1}{F^{2}}\iint L_c(u,v,x,y) A(u,v)\cos^{4}(\theta )\: \mathrm{d}u\mathrm{d}v,
\end{align}
where $A(u,v)$ is the aperture function (unity inside and zero outside the aperture). $\mathcal{I}_c(x,y)$ represents a 2-D image focused at a depth $F$, and the depth range corresponding to focused region is determined by the size of $A(u,v)$, where a larger aperture leads to a narrower depth range. The refocusing of the LF $L_c(u,v,x,y)$ is equivalent to computing the intensity $\overline{\mathcal{I}_c}(x',y')$ of a 2-D image with a shifted image plane $(x',y')$ located at a distance $\alpha F$ from the camera plane $uv$. Note that, when $\alpha>1$, the LF is refocused to distant depths than the original focused depth, and when $\alpha<1$, the LF is refocused to near depths than the original focused depth. The intensity $\overline{\mathcal{I}_c}(x',y')$ of the refocused 2-D image can be computed as~\cite{ng2005light}
\begin{align}
\label{eq:LF_refocus}
\overline{\mathcal{I}_c}\left( x',y'\right) =\iint L_c\left( u,v,u+\frac{x' -u}{\alpha } ,v+\frac{y' -v}{\alpha }\right) \: \mathrm{d}u\mathrm{d}v,
\end{align}
where, the constant scaling factor $\frac{1}{F^{2}}$ is ignored, and integration is limited $A(u,v)=1$. Note that, the pixels of an SAI of the LF $L( u_0,v_0,x,y)$ at $(u,v)=(u_0,v_0)$ need to be shifted by $(u_0+\frac{x' -u_0}{\alpha})-x'$ and $(v_0+\frac{y' -v_0}{\alpha})-y'$ pixels before integrating with respect to to $(x',y')$, and the shifts are determined by $\alpha$ that depends on the depth of the refocused plane. We now consider a discrete-domain LF $L(n_u,n_v,n_x,n_y)=L_c(n_u\Delta u,n_v\Delta v,n_x\Delta x,n_y\Delta y)$, where $(n_u,n_v,n_x,n_y)\in\mathbb{Z}^4$ and $\Delta i$ ($i=u,v,x,y$) is the sampling interval corresponding to the dimension $i$. In this case, $\overline{\mathcal{I}}\left( n_x' ,n_y'\right)$  can be expressed for a dense LF consisting of $N_u\times N_v$ SAIs, from~\eqref{eq:LF_refocus}, as~\cite{ng2005light}
\begin{align}
\label{eq:LF_refocus_discrete}
\overline{\mathcal{I}}\left( n_x' ,n_y'\right) =&\sum_{n_u=1}^{N_u}\sum_{n_v=1}^{N_v} L\left( n_u,n_v,\frac{n_x'+(\alpha-1) n_u}{\alpha}, \right. \notag \\ 
& \left. \hspace{3cm} \frac{n_y'+(\alpha-1)n_v}{\alpha}\right).
\end{align}
Note that the factor $1/\alpha$ can be ignored because it just scale the refocused LF. This simplifies \eqref{eq:LF_refocus_discrete} to   
\begin{align}
\label{eq:LF_refocus_simple}
\overline{\mathcal{I}}\left( n_x' ,n_y'\right) =&\sum_{n_u=1}^{N_u}\sum_{n_v=1}^{N_v} L\left( n_u,n_v,n_x'+(\alpha-1) n_u, \right. \notag \\ 
& \left. \hspace{3cm} n_y'+(\alpha-1)n_v\right),
\end{align}
where refocused 2-D image is generated by \emph{shifting and summing} the SAIs of the LF. 


\section{Proposed Multi Arbitrary-Volume Refocusing Algorithm} \label{sec:methodology}

\begin{figure*}[!ht]
    \centering
    \begin{subfigure}[c]{0.5\textwidth}
         \centering
         \includegraphics[width=\textwidth]{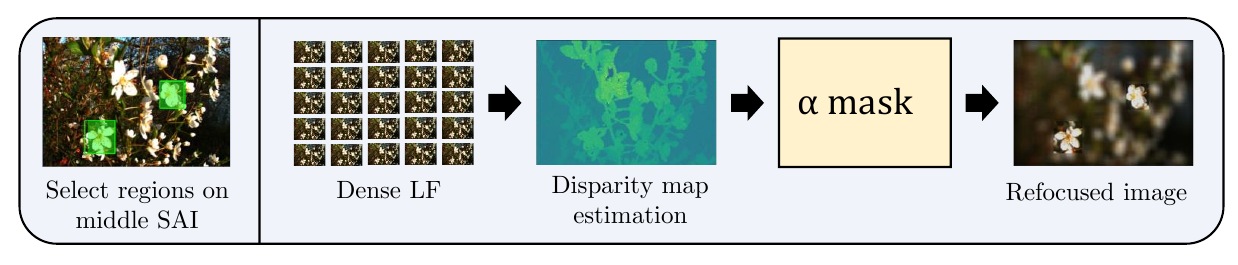}
         \caption{ }
         \label{fig:dense_refocus_pipeline}
     \end{subfigure}
     \hfill
    \begin{subfigure}[c]{1\textwidth}
         \centering
         \includegraphics[width=0.8\textwidth]{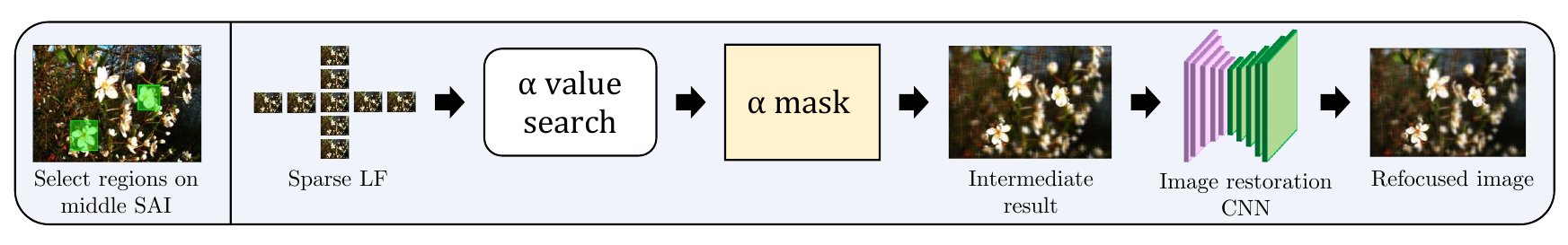}
         \caption{ }
         \label{fig:sparse_refocus_pipeline}
     \end{subfigure}
     \hfill
    
    \caption{The flow charts of (a) dense LF refocusing algorithm and (b) sparse LF refocusing algorithm. The user can interactively select an ROI or multiple ROIs on middle SAI which are needed to be refocused. For dense LFs, first, the depth map is estimated. Then generate the $\mathcal{M}_{\alpha}(n_x,n_y)$. Finally refocus each region with different $\alpha$ value separately and concatenate together to get the final refocused image. For sparse LFs, first suitable $\alpha$ value search is done for each ROI. Then create the $\mathcal{M}_{\alpha}(n_x,n_y)$ and refocus each region with different $\alpha$ value and concatenate. Finally, send through the image restoration CNN to remove the aliasing artifacts and get the final refocused image.}
    \label{fig:architecture}
\end{figure*}

In this section, we propose algorithms for interactive refocusing of arbitrarily selected $N$ $(\geq 1)$ regions of interest (ROIs) of dense LFs and cross-shaped sparse LFs. Both algorithms are initiated by the user by providing the required ROIs to be refocused on the middle SAI and their visible depth ranges. Let's denote the selected $N$ ROIs as $\mathcal{R}^1, \mathcal{R}^2, ..., \mathcal{R}^N$ and their visible depth ranges as $\phi^1, \phi^2, ..., \phi^N$ where each $\phi^i$ denotes whether $\mathcal{R}^i \in \mathfrak{R}_{n}$ (narrow-depth) or $\mathcal{R}^i \in \mathfrak{R}_{w}$ (wide-depth). This prior information improves the efficiency of the algorithms as the amount of calculations for $\mathcal{R}^i \in \mathfrak{R}_{n}$ is relatively low.

\subsection{Refocusing of Dense Light Fields}\label{sec:dense_refocus}

\begin{figure}[!h]
    \centering
    \includegraphics[width=0.9\columnwidth]{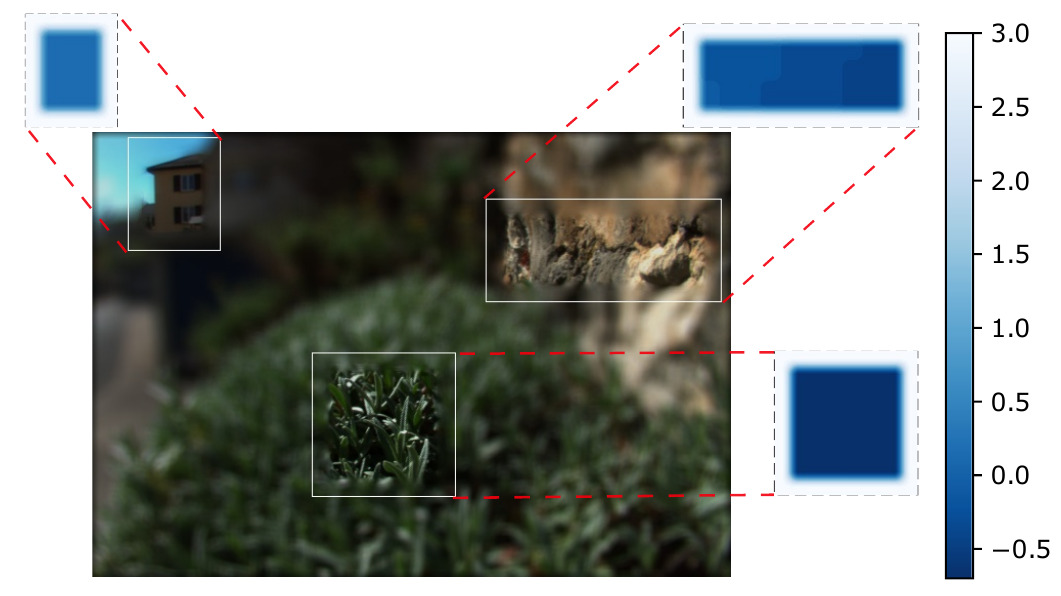}
    \caption{Sections of $\mathcal{M}_{\alpha}(n_x,n_y)$ of multi arbitrary-volume refocused LF shown in \figurename~\ref{fig:dense_multi_refocused_img_slices} as a heatmap.}
    \label{fig:alpha_mask}
\end{figure}

The proposed refocusing algorithm for dense LFs consists of 2 steps. \figurename~\ref{fig:dense_refocus_pipeline} shows the overall pipeline of the process. First, the disparity map $D(n_x,n_y)$ is estimated. To this end, we utilize fast implementable, LF epipolar plane image analysis in~\cite{wanner2013variational} to generate a smooth $D(n_x,n_y)$. Here, we first estimate gradients of epipolar lines $d_{x,u}$ and $d_{y,v}$ passing through  $(n_u,n_x)$ and $(n_v,n_y)$ points, respectively, and their corresponding confidences. We then point-wise select the most confident estimate among them and apply smoothing to attenuate noise. Incorporating $D(n_x,n_y)$, we can find $\alpha$ values to refocus each $(n_x,n_y)$ pixel of the LF as 
\begin{align}\label{eq:disparity}
\alpha(n_x,n_y) = D(n_x,n_y) -1,
\end{align}
where $\alpha(n_x,n_y)$ denotes the corresponding $\alpha$ value for the pixel $(n_x,n_y)$. Then, for each ROI $\in \mathfrak{R}_{n}$, we select the mode of $\alpha$ values as the suitable $\alpha$ value in~\eqref{eq:LF_refocus_simple} to refocus that ROI, whereas each ROI $\in \mathfrak{R}_{w}$ are divided into smaller patches $(p^1, p^2, ..., p^M)$ and find the mode of $\alpha$ values for each patch. To emphasize the selected in-focus regions, we set a relatively smaller $\alpha$ value on uninterested regions which makes those regions out-of-focus. Concatenating these $\alpha$ values, we create the $\alpha$ mask $\mathcal{M}_{\alpha}(n_x,n_y)$. In order to ensure a smooth transition between regions with different $\alpha$ values, we apply a $15 \times 15$ Gaussian Filter with $\mu=0$ and $\sigma=5$ on $\mathcal{M}_{\alpha}(n_x,n_y)$ and remove sharp changes. \figurename~\ref{fig:alpha_mask} shows sections of $\mathcal{M}_{\alpha}(n_x,n_y)$ around the ROIs as heat map parts. Note that the smooth transitions between focused and out-of-focus regions, and continuous change of $\alpha$ value within ROI $\in \mathfrak{R}_{w}$ (upper right ROI). Using the constructed $\mathcal{M}_{\alpha}(n_x,n_y)$, we identify the regions with unique $\alpha$ values, then separately refocus using~\eqref{eq:LF_refocus_simple} and concatenate together to get the final multi arbitrary-volume refocused image $\mathcal{I}_f$ (\figurename~\ref{fig:dense_multi_refocused_img_slices}). The overall algorithm for dense LF refocusing is presented in Algorithm \ref{alg:dense_LF_refocus}. Note that, though we can find $\alpha$ value and refocus each pixel individually, we find a single $\alpha$ value for whole ROI $(\in \mathfrak{R}_{n})$ or patch $(\in \mathfrak{R}_{w})$ and refocus at once, as in such regions $\alpha$ values have less variances. So that the computational efficiency can be further improved.

\begin{algorithm}[!h]
\scriptsize
\caption{Dense LF multi arbitrary-volume refocusing}\label{alg:dense_LF_refocus}

\hspace*{\algorithmicindent} \textbf{Input:} Original Light Field ($\mathcal{I}_{LF}$) \\
\hspace*{\algorithmicindent} \hspace{.9cm} Default $\alpha$ ($\alpha_d$)\\ 
\hspace*{\algorithmicindent} \textbf{Output:} Refocused Light Field ($\mathcal{I}_f$)
\begin{algorithmic}[1]
\Procedure{Dense\_LF\_Multi\_Arbitrary\_Volume\_Refocus}{$\alpha_{d}$, $\mathcal{I}_{LF}$}
    \State $\mathcal{R}_{ROI} \gets  [\mathcal{R}^1, \mathcal{R}^2, ..., \mathcal{R}^N]$
    \Comment{User selected ROIs}
    \State $\Phi \gets [\phi^1, \phi^2, ..., \phi^N],~\phi^i \in \{\mathfrak{R}_{n}, \mathfrak{R}_{w}\}$
    \Comment{Mode of refocus}
    \State $\mathcal{I}_f \gets \{0:\forall(n_x,n_y)\}$ \Comment{Initialize $\mathcal{I}_f$ to zeros}
    \State $\mathcal{M}_{\alpha}$ $\gets$ \Call{Generate\_Alpha\_Mask}{$\alpha_{d}$, $\mathcal{R}_{ROI}$, $\Phi$, $\mathcal{I}_{LF}$};
    
    \ForEach{$\alpha$ in $\mathcal{M}_\alpha$}
        \State $I_{\alpha}$ $\gets$ \Call{LF\_Refocus}{$\mathcal{I}_{LF}$, $\alpha$} \Comment{~\eqref{eq:LF_refocus_simple}}
        \State $\mathcal{K} \gets \{(n_x,n_y):\mathcal{M}_{{\alpha}_{(n_x,n_y)}}=\alpha\}$
        \State $\mathcal{I}_f[n_x,n_y]$ $\gets$ $I_{\alpha}[n_x,n_y] \;\forall(n_x,n_y)\in\mathcal{K}$

    \EndFor
    \State \textbf{return} $\mathcal{I}_f$
\EndProcedure



\Procedure{Generate\_Alpha\_Mask}{$\alpha_{d}$, $\mathcal{R}_{ROI}$, $\Phi$, $\mathcal{I}_{LF}$}
    \State Generate disparity map $D(n_x,n_y)$
    \State $D(n_x,n_y) \gets D(n_x,n_y) + 1$ \Comment{~\eqref{eq:disparity}}
    \State $\mathcal{M}_\alpha \gets \{\alpha_d:\forall(n_x,n_y)\}$ \Comment{Initialize $\mathcal{M}_\alpha$ to $\alpha_d$} 
    \For{$i \gets 1$ to $N$}
    \State $\mathcal{R}_{c} \gets \mathcal{R}_{ROI}[i]$
    \State $\phi_{c} \gets \Phi[i]$
        \If{($\phi_c == \mathfrak{R}_{n}$)}
            \State $\alpha$ $\gets$ mode($\alpha$ $\in$ $D$[$\mathcal{R}_c$])
            \State $\mathcal{M}_\alpha$[$\mathcal{R}_c$] $\gets$ $\alpha$
        \Else  \Comment{$\phi_c \in \mathfrak{R}_{w}$}
        
        \Comment{Divide $\mathcal{R}_c$ into smaller patches $p^j$, $j=1,2,..,M$}
        \State $\mathcal{P}_c \gets$ $[p^1, p^2, ..., p^M]$ 
        \For{$j \gets 1$ to $M$}  
        \State $p_c \gets$ $\mathcal{P}_c[j]$
        \Comment{$j^{th}$ patch in $\mathcal{R}_c$}

                \State $\alpha$ $\gets$ mode($\alpha$ $\in$ $D$[$p_c$])
                \State $\mathcal{M}_\alpha$[$p_c$] $\gets$ $\alpha$
            \EndFor
        \EndIf
    \EndFor
    \State Apply Gaussian filter on $\mathcal{M}_\alpha$
    \State \textbf{return} $\mathcal{M}_\alpha$
\EndProcedure

\end{algorithmic}
\end{algorithm}

\subsection{Sparse Light Field Refocusing}\label{sec:sparse_refocus}

In this paper, we consider a cross-shaped SAIs of a dense LF as a sparse LF. The structure of the EPIModule sparse LF camera~(\figurename~\ref{fig:epiModule}) motivated us to consider this specific shape.
So that, our sparse LF refocusing algorithm can be directly implemented using sparse LFs acquired from this kind of cameras, without depending on much expensive dense LFs.

Our proposed sparse LF post-capture refocusing algorithm consists of 3 steps. \figurename~\ref{fig:sparse_refocus_pipeline} shows the overall pipeline. First, for each ROI $\in \mathfrak{R}_{n}$, a suitable $\alpha$ value which can be used to refocus that region is searched. ROIs $\in \mathfrak{R}_{w}$ are divided into smaller patches $(p^1, p^2, ..., p^M)$ and find a suitable $\alpha$ value for each patch separately. For this $\alpha$ value search, we extract the corresponding LF slice $(\mathcal{S}_{LF})$ of an ROI or patch and refocus with different $\alpha$ values using~\eqref{eq:LF_refocus_simple}. We measure the similarity between resulting refocused image slices and the corresponding area of the LF's middle SAI $(\mathcal{S}_{SAI})$ using \emph{multi scale structural similarity} (MS\_SSIM)~\cite{ms_ssim} metric. The $\alpha$ value which gives the highest similarity is selected as the best $\alpha$ value. This method is repeatedly applied for all the ROIs or patches to identify the set of best $\alpha$ values. This is done in two steps. In the first step, a range that the best $\alpha$ value exists is identified using the fact that the similarity score has a concave shape and global maxima is at the best $\alpha$ value. Then, the best $\alpha$ value is identified by exhaustively searching within identified range.

Following the dense LF refocusing algorithm, we create a $\mathcal{M}_{\alpha}(n_x,n_y)$ consists of corresponding $\alpha$ values for each region. After that, each region with an unique $\alpha$ value is separately refocused using~\eqref{eq:LF_refocus_simple} and concatenate together to get the intermediate multi arbitrary-volume refocused image $\mathcal{I}_t$. But this result suffers from ghosting artifacts in out-of-focus regions due to the undersampling of LF. To reduce this artifact, we post-process  $\mathcal{I}_t$ using an image restoration CNN to get the final refocused image  $\mathcal{I}_f$ (\figurename~\ref{fig:sparse_multi_refocused_img_slices}). The overall algorithm for sparse LF refocusing is presented in Algorithm \ref{alg:sparse_LF_refocus}. The proposed aliasing reduction method is discussed in the following section \ref{sec:cnn}.

\begin{algorithm}[!h]
\scriptsize
\caption{Sparse LF multi arbitrary-volume refocusing}\label{alg:sparse_LF_refocus}

\hspace*{\algorithmicindent} \textbf{Input:} Original Light Field ($\mathcal{I}_{LF}$) \\
\hspace*{\algorithmicindent} \hspace{.9cm} Default $\alpha$ ($\alpha_d$)\\ 
\hspace*{\algorithmicindent} \textbf{Output:} Refocussed Light Field ($\mathcal{I}_f$)

\begin{algorithmic}[1]

\Procedure{Sparse\_LF\_Multi\_Volume\_Refocus}{$\alpha_{d}$, $\mathcal{I}_{LF}$}
\State $\mathcal{R}_{ROI} \gets  [\mathcal{R}^1, \mathcal{R}^2, ..., \mathcal{R}^N]$
    \Comment{User selected ROIs}
    \State $\Phi \gets [\phi^1, \phi^2, ..., \phi^N],~\phi^i \in \{\mathfrak{R}_{n}, \mathfrak{R}_{w}\}$
    \Comment{Mode of refocus}
    \State $\mathcal{I}_t \gets \{0:\forall(n_x,n_y)\}$ \Comment{Initialize $\mathcal{I}_t$ to zeros}
    \State $\mathcal{M}_\alpha$ $\gets$ \Call{Generate\_Alpha\_Mask}{$\alpha_{d}$, $\mathcal{R}_{ROI}$, $\Phi$, $\mathcal{I}_{LF}$}

    \ForEach{$\alpha$ in $\mathcal{M}_\alpha$}
        \State $I_{\alpha}$ $\gets$ \Call{LF\_Refocus}{$\mathcal{I}_{LF}$, $\alpha$} \Comment{Eq.~\ref{eq:LF_refocus_simple}}
        \State $\mathcal{K} \gets \{(n_x,n_y):\mathcal{M}_{{\alpha}_{(n_x,n_y)}}=\alpha\}$
        \State $\mathcal{I}_f[n_x,n_y]$ $\gets$ $I_{\alpha}[n_x,n_y] \;\forall(n_x,n_y)\in\mathcal{K}$;

    \EndFor
    \State $\mathcal{I}_f \gets$ \Call{Remove\_Aliasing\_Artifacts}{$\mathcal{I}_t$}
    \State \textbf{return} $\mathcal{I}_f$
\EndProcedure



\Procedure{Generate\_Alpha\_Mask}{$\alpha_{d}$, $\mathcal{R}_{ROI}$, $\Phi$, $\mathcal{I}_{LF}$}
    \State $\mathcal{M}_\alpha \gets \{\alpha_d:\forall(n_x,n_y)\}$ \Comment{Initialize $\mathcal{M}_\alpha$ to $\alpha_d$} 
    \For{$i \gets 1$ to $N$}
        \State $\mathcal{R}_{c} \gets \mathcal{R}_{ROI}[i]$
        \State $\phi_{c} \gets \Phi[i]$
        \If{($\phi_c == \mathfrak{R}_{n}$)}
            \State $\mathcal{S}_{LF} \gets Slice(\mathcal{I}_{LF},\mathcal{R}_c) $;
            \State $\mathcal{S}_{SAI} \gets Slice(\mathcal{I}_{SAI},\mathcal{R}_c) $
            \State $[\alpha_{min},\alpha_{max}]$ $\gets$ \Call{Get\_Alpha\_Range}{$\mathcal{S}_{LF}$, $\mathcal{S}_{SAI}$}
            \State $\alpha$ $\gets$ \Call{Get\_Alpha}{$\mathcal{S}_{LF}$, $\mathcal{S}_{SAI}$, $\alpha_{min}$, $\alpha_{max}$};
            \State $\mathcal{M}_\alpha$[$\mathcal{R}_c$] $\gets$ $\alpha$
        \Else  \Comment{$\mathcal{R}_c \in \mathfrak{R}_{w}$, Divide $\mathcal{R}_c$ into smaller patches $p^j$, $j=1,2,..,M$}

        \State $\mathcal{P}_c \gets$ $[p^1, p^2, ..., p^M]$ 
        \For{$j \gets 1$ to $M$}  
        \State $p_c \gets$ $\mathcal{P}_c[j]$
        \Comment{$j^{th}$ patch in $\mathcal{R}_c$}
        \State ${\mathcal{S}_{LF}} \gets Slice(\mathcal{I}_{LF}, p_c) $
        \State ${\mathcal{S}_{SAI}} \gets Slice(\mathcal{I}_{SAI}, p_c) $
        \State $[\alpha_{min},\alpha_{max}]$ $\gets$ \Call{Get\_Alpha\_Range}{${\mathcal{S}_{LF}}$, ${\mathcal{S}_{SAI}}$}
                \State $\alpha$ $\gets$ \Call{Get\_Alpha}{${\mathcal{S}_{LF}}$, ${\mathcal{S}_{SAI}}$, $\alpha_{min}$, $\alpha_{max}$}
                \State $\mathcal{M}_\alpha$[$p_c$] $\gets$ $\alpha$;
            \EndFor
        \EndIf
    \EndFor
    \State \textbf{return} $\mathcal{M}_\alpha$
\EndProcedure



\Procedure{Get\_Alpha\_Range}{$\mathcal{I}^S_{LF}$, $\mathcal{I}^S_{SAI}$}
    \State $\alpha_{l} \gets 0.0$; $\alpha_{m} \gets 1.0$; $\alpha_{r} \gets 2.0$; \Comment{Initialize}
    
    \While {True}
        \State $\mathcal{I}_{l}$ $\gets$ \Call{LF\_Refocus}{$\mathcal{I}^S_{LF}$, $\alpha_{l}$} \Comment{Eq.~\ref{eq:LF_refocus_simple}}
        \State $\mathcal{I}_{m}$ $\gets$ \Call{LF\_Refocus}{$\mathcal{I}^S_{LF}$, $\alpha_{m}$} \Comment{Eq.~\ref{eq:LF_refocus_simple}}
        \State $\mathcal{I}_{r}$ $\gets$ \Call{LF\_Refocus}{$\mathcal{I}^S_{LF}$, $\alpha_{r}$} \Comment{Eq.~\ref{eq:LF_refocus_simple}}
        
        \State $SIM_{\alpha_{l}}$ $\gets$ \Call{MS\_SSIM}{$\mathcal{I}_{l}$, $\mathcal{I}^S_{SAI}$}
        \State $SIM_{\alpha_{m}}$ $\gets$ \Call{MS\_SSIM}{$\mathcal{I}_{m}$, $\mathcal{I}^S_{SAI}$}
        \State $SIM_{\alpha_{r}}$ $\gets$ \Call{MS\_SSIM}{$\mathcal{I}_{r}$, $\mathcal{I}^S_{SAI}$}
        \vspace{.1cm}
        \If{($SIM_{\alpha_{l}}<SIM_{\alpha_{m}}$ \& $SIM_{\alpha_{m}} < SIM_{\alpha_{r}}$)}
            \State $\alpha_{l}\gets\alpha_{l}+1; \alpha_{m}\gets\alpha_{m}+1; \alpha_{r}\gets\alpha_{r}+1;$
        
        \ElsIf {($SIM_{\alpha_{l}}>SIM_{\alpha_{m}}$ \& $SIM_{\alpha_{m}}>SIM_{\alpha_{r}}$)} 
            \State $\alpha_{l}\gets\alpha_{l}-1; \alpha_{m}\gets\alpha_{m}-1; \alpha_{r}\gets\alpha_{r}-1;$
        \Else  \Comment{($SIM_{\alpha_{l}} < SIM_{\alpha_{m}}$ \& $SIM_{\alpha_{r}} > SIM_{\alpha_{m}}$)}
            \State \textbf{return} [$\alpha_{l},\alpha_{r}$]
        \EndIf
    \EndWhile
\EndProcedure



\Procedure{Get\_Alpha}{$\mathcal{I}^S_{LF}$, $\mathcal{I}^S_{SAI}$,  $\alpha_{min}$, $\alpha_{max}$}
    \State $SIM_{prev} \gets 0.0$; $\alpha_{best} \gets 0.0$; \Comment{Initialize}
    \For{($\alpha \gets \alpha_{min}$ : $\alpha \leq \alpha_{max}$ : $\alpha \mathrel{{+}{=}} \Delta\alpha$)}
        \State $\overline{\mathcal{I}_t}$ $\gets$ \Call{LF\_Refocus}{$\mathcal{I}^S_{LF}$, $\alpha$} \Comment{Eq.~\ref{eq:LF_refocus_simple}}
        \State $SIM_{\alpha}$ $\gets$ \Call{MS\_SSIM}{$\overline{\mathcal{I}_t}$, $\mathcal{I}^S_{SAI}$}
        \If{$SIM_{\alpha}\geq SIM_{prev}$}
            \State $\alpha_{best} \gets \alpha$
            \State $SIM_{prev} \gets SIM_{\alpha}$
        \EndIf
    \EndFor
    \State \textbf{return} $\alpha_{best}$
\EndProcedure
\end{algorithmic}
\end{algorithm}

\subsection{Reducing Aliasing Artifacts using Image Restoration CNN} \label{sec:cnn}

\begin{figure*}[!t]
    \centering

    \begin{subfigure}[c]{0.64\columnwidth}
         \centering
         \includegraphics[width=\textwidth]{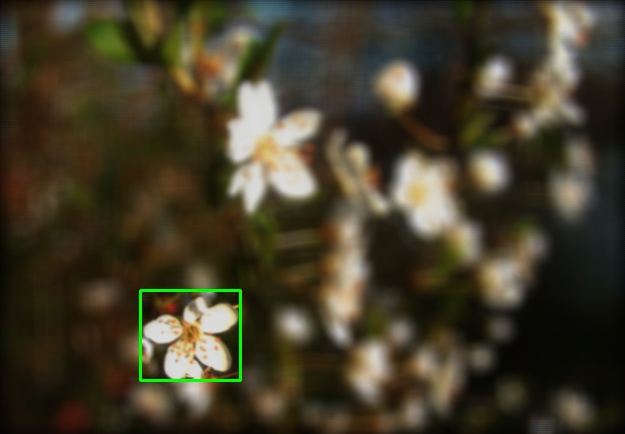}
         \caption{}
         \label{fig:dense_refocused_img}
     \end{subfigure}
     \hfill
    \begin{subfigure}[c]{0.64\columnwidth}
         \centering
         \includegraphics[width=\textwidth]{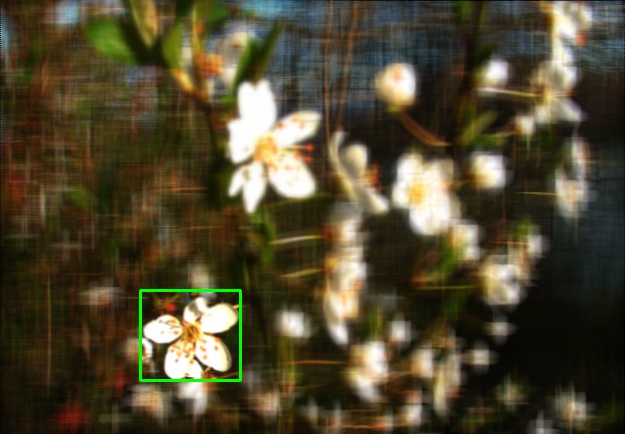}
         \caption{}
         \label{fig:sparse_before_cnn}
     \end{subfigure}
     \hfill
     \begin{subfigure}[c]{0.64\columnwidth}
         \centering
         \includegraphics[width=\textwidth]{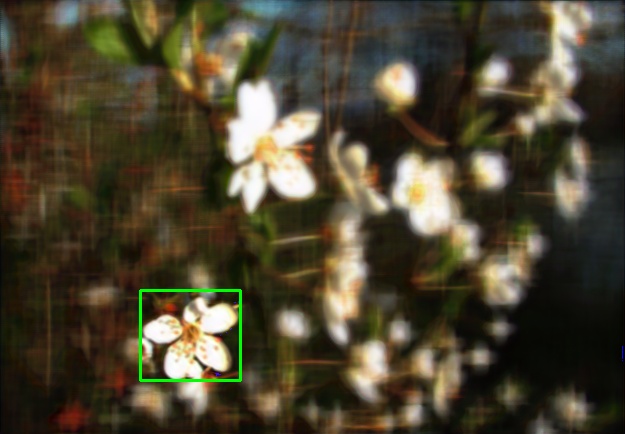}
         \caption{}
         \label{fig:sparse_after_cnn}
     \end{subfigure}
    \caption{Refocused image quality comparison between dense and sparse LF refocusing methods; (a) refocused using the dense LF; (b) refocused cross-shaped sparse LF without post-processing; (c) refocused using cross-shaped sparse LF with post-processing.}
    \label{fig:denseSparsPostProcess}
\end{figure*}

The ghosting artifacts due to undersampling of LF makes $\mathcal{I}_t$ image unnatural. Therefore, these artifacts should be minimized. To this end, we approach this problem as an image restoration task. The  $\mathcal{I}_t$ is post-processed using an image restoration CNN to remove the ghosting artefacts and further improve the quality and generate $\mathcal{I}_f$. As an example \figurename~\ref{fig:dense_refocused_img} shows the ``Mirabelle Prune Tree" dense LF from EPFL LF dataset~\cite{epfldataset} refocused using~\eqref{eq:LF_refocus_simple} for single ROI. It has been properly refocused to the required depth while out-of-focus regions are blurred. \figurename~\ref{fig:sparse_before_cnn} shows the same refocused image but using the corresponding sparse LF. Due to the undersampling, this image has clearly visible ghosting artifact in out-of-focus regions. After post-processing this refocused image using the image restoration CNN, ghosting artifact is reduced significantly as shown in \figurename~\ref{fig:sparse_after_cnn}. 


Out of the many state-of-the-art data-driven methodologies~\cite{hinet,mprnet} that perform the image restoration have been derived from U-Net~\cite{unet}, a base network which is used predominantly for medical image restoration. Modern image restoration algorithms~\cite{swinir} have shifted towards the adoption of vision transformer-based (ViT) principles, however, with the requirement of vast amount of data to learn the restoration patterns. This is, however, a bottleneck for us as the number of LFs available are too little to learn a robust image restoration ViT. Therefore, we adopt U-Net~\cite{unet} architecture with some modifications for our image restoration CNN.

\subsubsection{Ground Truth and Input Generation}
The inputs to the image restoration network are generated using our proposed sparse LF refocusing algorithm that is discussed in Section \ref{sec:sparse_refocus}. To perform supervised training to remove the aliasing artefacts present in the inputs, a corresponding ground truth is generated using our proposed dense LF refocusing algorithm that is discussed in Section \ref{sec:dense_refocus}. 


To address the lack of data to learn the general patterns to restore a given  $\mathcal{I}_t$ refocused image with ghosting effects present, we utilize patch-wise training. Following \cite{deep_sparse}, the patches are generated by cropping $m \times m$ areas of the input in a non-overlapping manner. In our method, we used $m=100$ to reduce the computational complexity required to run through an entire input. 

\subsubsection{Loss Function}

Modifying the traditional loss function for image restoration, we propose a loss function $\mathcal{L}$ that combines conventional \emph{mean squared error loss} ($\mathcal{L}_{MSE}$), \emph{$L_1$ loss} ($\mathcal{L}_{L_1}$) with MS\_SSIM~\cite{ms_ssim} and PSNR~\cite{psnr} to preserve image quality. The loss function $\mathcal{L}$ is defined as   
\begin{align} \label{loss_func}
    \mathcal{L} = \mathcal{L}_{MSE} + \beta\mathcal{L}_{SSIM} + (1 - \beta)\mathcal{L}_{L_1} + \gamma\mathcal{L}_{PSNR},
\end{align}
where
\begin{align}
\mathcal{L}_{SSIM} &= 1-\mathrm{MS\_SSIM}( P_{g} ,P_{p}) \\
\mathcal{L}_{PSNR} &= 1/\mathrm{PSNR}( P_{g} ,P_{p}),
\end{align}
Here, $P_{g}$ and $P_{p}$ denote ground truth patch and predicted patch. $\beta, \gamma$ are the tuning parameters. Similar type of loss function is used in \cite{machinelearning_multi_refocus}. In our work, we utilize  0.65 and 500 for $\beta$ and $\gamma$, respectively.



\section{Experimental Results} \label{sec:experiments}

In this section, we discuss about our implementation specific details, qualitative and quantitative results of proposed algorithms. We provide detailed comparison between our two algorithms for dense and sparse LFs.

\subsection{Datasets}

We tested our proposed algorithms on LFs from stanford dataset~\cite{stanfordnewdataset}, HCI dataset~\cite{hcidataset} and EPFL dataset~\cite{epfldataset} to evaluate their performances.
For LFs in EPFL dataset~\cite{epfldataset} acquired using an LF camera, we remove the SAIs at the boarder of the grid, due to vignetting effect. 
The LF dimensions are shown in Table \ref{tab:Datasets}.

\begin{table}[H]
    \centering
    \caption{Dataset details}
    \begin{tabularx}{\linewidth}{ >{\raggedright\arraybackslash}m{2.0cm} |
    >{\centering\arraybackslash}m{2.0cm} |
    >{\centering\arraybackslash}m{2.0cm}
    }
        \noalign{\vskip 1.5pt}
        \hlineB{3}
        \noalign{\vskip 1.5pt}
        \textbf{Dataset} & \textbf{SAIs} & \textbf{Resolution} \\ 
        \noalign{\vskip 1.5pt}
        \hlineB{3}
        \noalign{\vskip 1.5pt}
        EPFL~\cite{epfldataset} & $15\times15$ & $434\times625$ \\
        HCI~\cite{hcidataset} & $9\times9$ & $512\times512$ \\
        Stanford~\cite{stanfordnewdataset} & $17\times17$ & $1024\times1024$ \\
        \noalign{\vskip 1.5pt}
        \hlineB{3}
    \end{tabularx}
    
    \label{tab:Datasets}
\end{table}

\subsection{Implementation Details} \label{sec:implementationDetails}

ROIs $\in \mathfrak{R}_{w}$ are divided into $20\times20$ pixels patches $(p^i)$ and find a suitable $\alpha$ value separately. The $\alpha$ value search for sparse LF refocusing algorithm finds the nearest $\alpha$ value to the first decimal point $(\Delta\alpha=0.1)$.
Our proposed algorithms are implemented using PyTorch to facilitate from GPU acceleration. Furthermore, the SAIs are shifted to the nearest pixel instead of shifting to sub-pixel level using interpolation to speed-up the process. We empirically found that the quality reduction in this approximation is negligible. 

The U-Net~\cite{unet} which is used for aliasing artifact removal in sparse LF image refocusing, is trained upto 30 epochs with a batch-size of 256. The weights are initialized with Xavier initialization~\cite{xavier} and used RMSProp optimizer. The learning rate is kept at 0.001.
As the network is fully convolutional, inferencing was done to whole image at once. The training and inferencing was done on Google Colaboratory (Tesla T4 GPU and Intel(R) Xeon(R) CPU @ 2.30GHz).

\subsection{Experimental Results}

\begin{table*}[!t]
    \centering
    \caption{Dense and sparse LF multi arbitrary-volume refocusing algorithm computational time and refocused images' BRISQUE~\cite{brisque} scores. All the time values are in seconds.}
    \begin{tabularx}{\textwidth}{m{2.5cm} c|c c c c >{\centering\arraybackslash}m{1.1cm}|c c c c c >{\centering\arraybackslash}m{0.8cm} }
        \hlineB{3}
      \noalign{\vskip 1.5pt} 
      \multirow{3}{*}{\textbf{Light Field}} & \multirow{3}{*}{$N$} & \multicolumn{5}{c|}{\textbf{Dense LF refocusing}} & \multicolumn{6}{c}{\textbf{Sparse LF refocussing}} \\ 
      \noalign{\vskip 1.5pt} 
      \cline{3-13}
      \noalign{\vskip 1.5pt} 
      & & $N^{d}_{\alpha}$ &  $T^{d}_{mask}$ & $T^{d}_{refocus}$ &  $T^{d}_{total}$ & BRISQUE value~\cite{brisque} & $N^{s}_{\alpha}$ &  $T^{s}_{mask}$ & $T^{s}_{refocus}$ & $T^{s}_{artifact}$ & $T^{s}_{total}$ & BRISQUE value~\cite{brisque}\\
      \noalign{\vskip 1.5pt} \hlineB{3}
      \noalign{\vskip 1.5pt}
      Lego Knights & 4 & 81 & 8.98 & 10.96 & 19.94 & 44.63 & 66 & 2.37 & 8.99 & 0.35 & 11.71 & 69.11 \\
      Mirablelle Prune Tree & 4 & 19 & 2.87 & 3.27 & 6.15 & 47.58 & 20 & 0.91 & 3.81 & 0.09 & 4.81 & 46.59 \\
      Books & 3 & 33 & 2.86 & 5.68 & 8.54 & 60.30 & 33 & 1.31 & 6.26 & 0.09 & 7.66 & 47.31 \\
      Sideboard & 3 & 34 & 3.23 & 0.40 & 3.64 & 14.96 & 34 & 0.29 & 0.35 & 0.08 & 0.71 & 42.83 \\ \hlineB{3}
    \end{tabularx}
    \label{tab:dense_sparse_resultls}
\end{table*}

To generate the results, several ROIs were selected on the middle SAI of LFs. Their coordinates and their visible depth ranges (whether the ROI  $\in \mathfrak{R}_{s}$ or $\in \mathfrak{R}_{w}$) were provided as mentioned in the section \ref{sec:methodology}. 

\begin{figure*}[!p]
    \centering
    
    \begin{subfigure}[c]{0.29\textwidth}
         \centering
         \caption*{Middle SAI}
     \end{subfigure}
     \hspace{0.1cm}
    \begin{subfigure}[c]{0.29\textwidth}
         \centering
         \caption*{\centering Dense LF multi arbitrary-volume refocused image}
     \end{subfigure}
     \hspace{0.1cm}
     \begin{subfigure}[c]{0.29\textwidth}
         \centering
         \caption*{\centering Sparse LF multi arbitrary-volume refocused image}
     \end{subfigure}

    \begin{subfigure}[t]{0.29\textwidth}
         \centering
         \includegraphics[width=\textwidth]{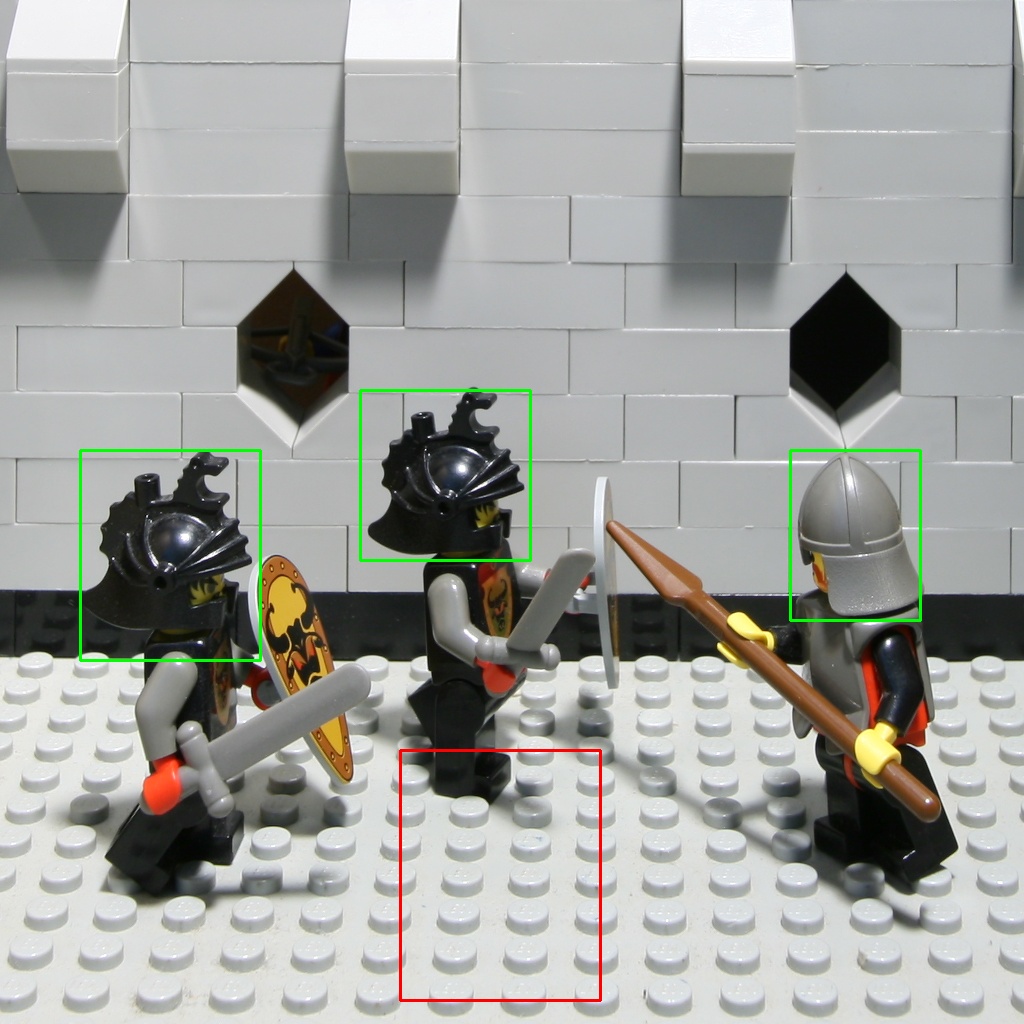}
     \end{subfigure}
     \hspace{0.1cm}
    \begin{subfigure}[t]{0.29\textwidth}
         \centering
         \includegraphics[width=\textwidth]{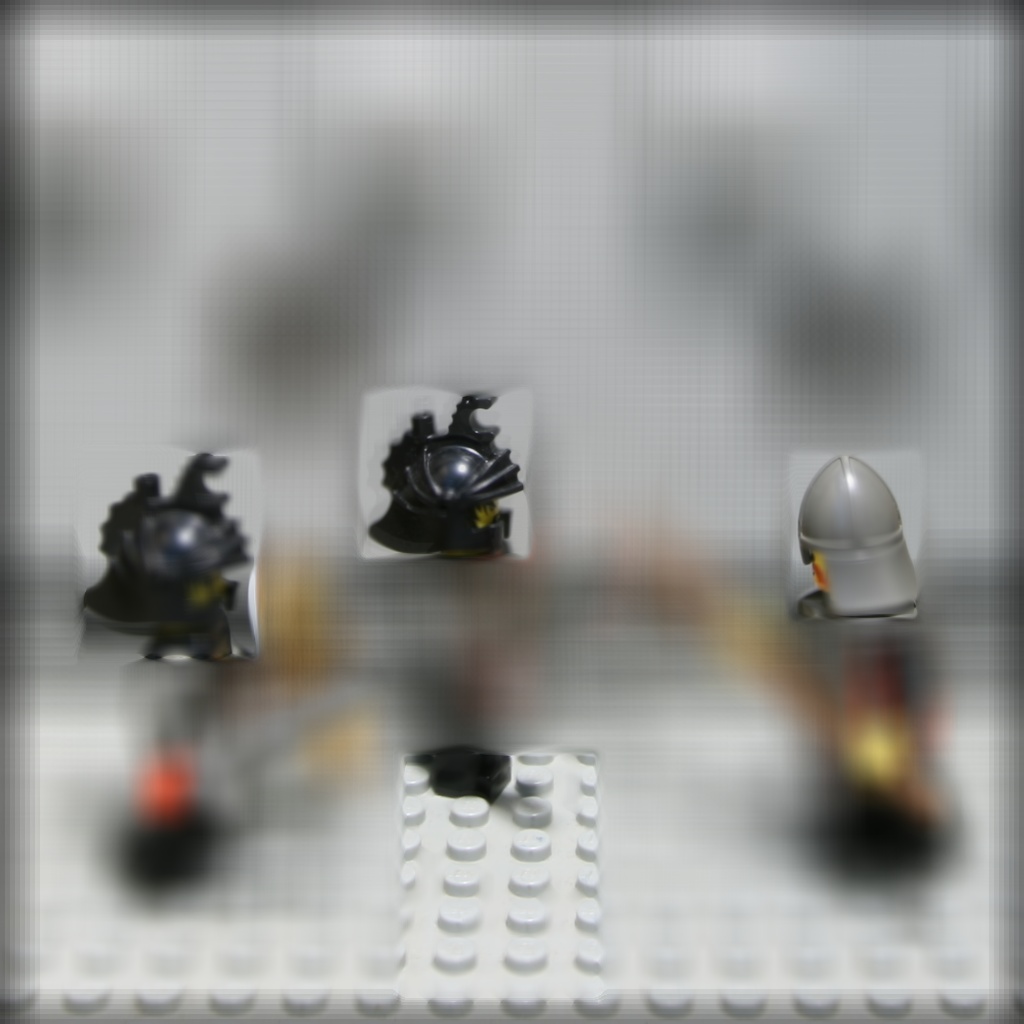}
         \caption{}
     \end{subfigure}
     \hspace{0.1cm}
     \begin{subfigure}[t]{0.29\textwidth}
         \centering
         \includegraphics[width=\textwidth]{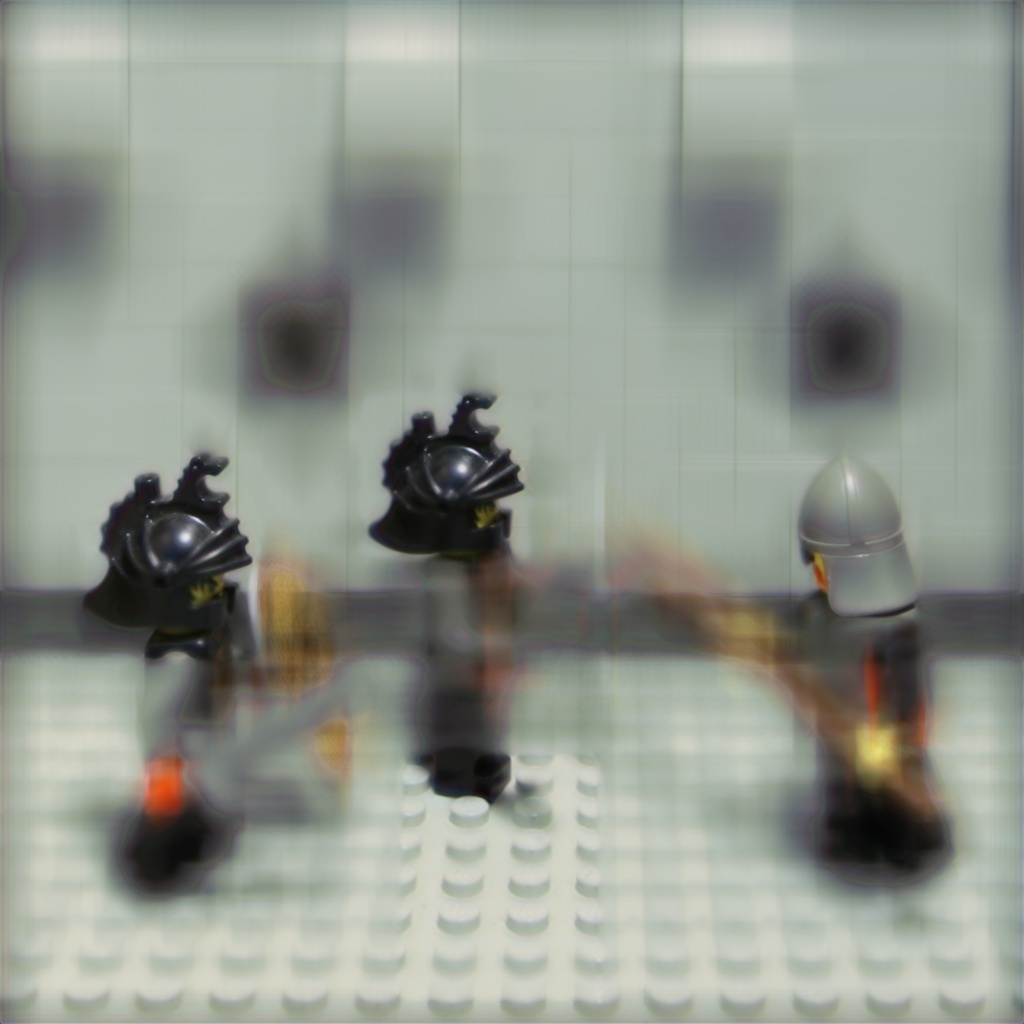}
     \end{subfigure}
     
    \vspace{0.2cm}

     
    \begin{subfigure}[t]{0.29\textwidth}
         \centering
         \includegraphics[width=\textwidth]{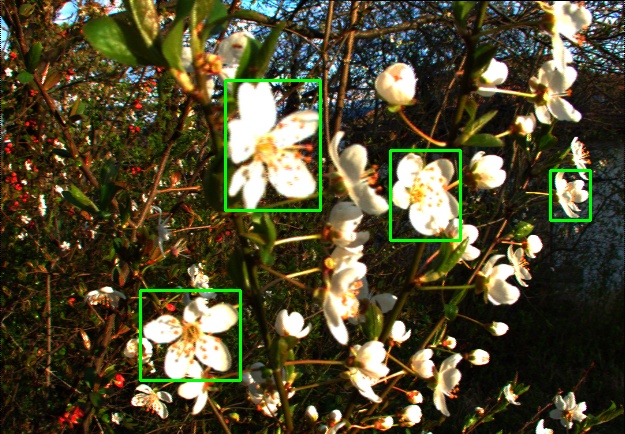}
     \end{subfigure}
     \hspace{0.1cm}
    \begin{subfigure}[t]{0.29\textwidth}
         \centering
         \includegraphics[width=\textwidth]{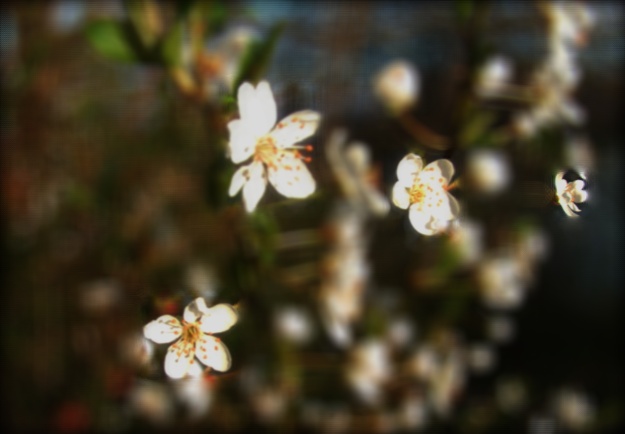}
         \caption{}
         \label{fig:mirabelle_prune_tree}
     \end{subfigure}
     \hspace{0.1cm}
     \begin{subfigure}[t]{0.29\textwidth}
         \centering
         \includegraphics[width=\textwidth]{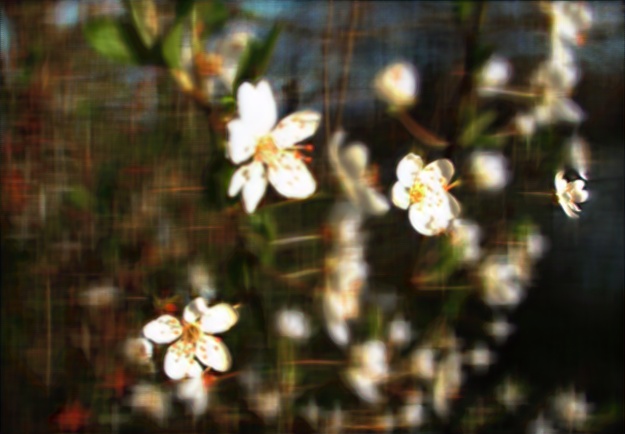}
     \end{subfigure}
    \vspace{.2cm}
     
    \begin{subfigure}[t]{0.29\textwidth}
         \centering
         \includegraphics[width=\textwidth]{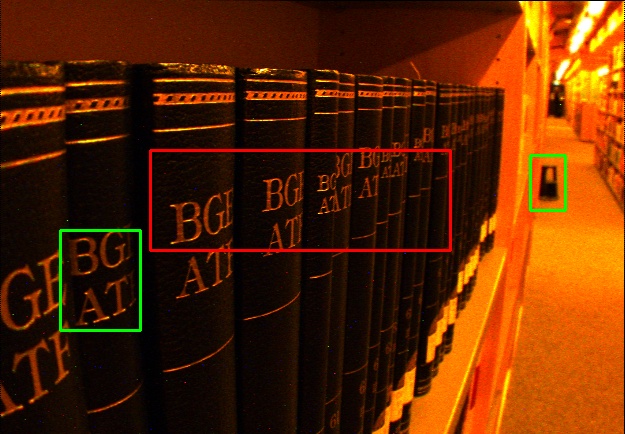}
     \end{subfigure}
     \hspace{0.1cm}
    \begin{subfigure}[t]{0.29\textwidth}
         \centering
         \includegraphics[width=\textwidth]{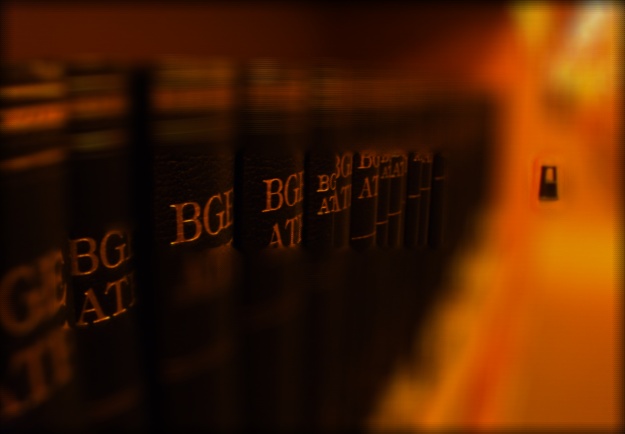}
         \caption{}
     \end{subfigure}
     \hspace{0.1cm}
     \begin{subfigure}[t]{0.29\textwidth}
         \centering
         \includegraphics[width=\textwidth]{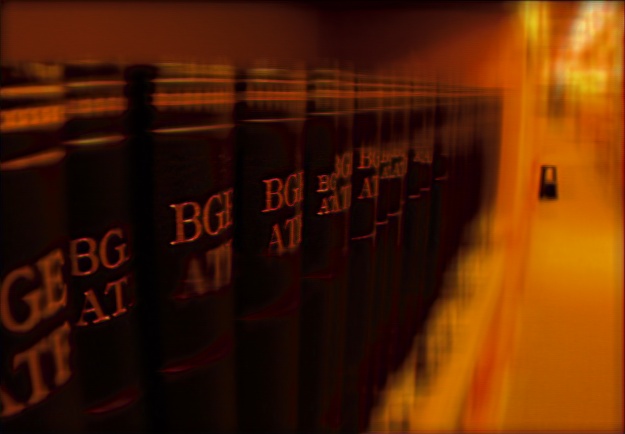}
     \end{subfigure}
     
    \vspace{.2cm}
     
    \begin{subfigure}[t]{0.29\textwidth}
         \centering
         \includegraphics[width=\textwidth]{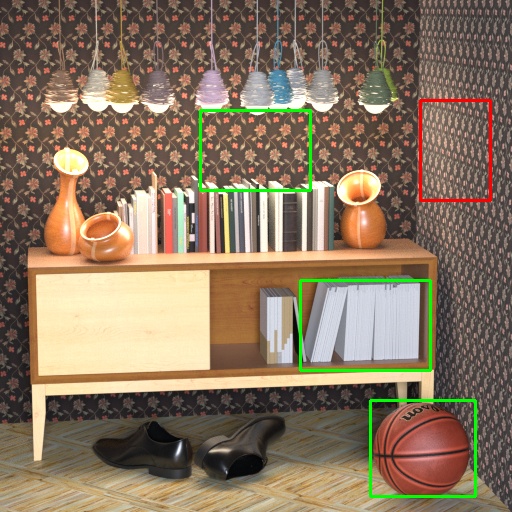}
     \end{subfigure}
     \hspace{0.1cm}
    \begin{subfigure}[t]{0.29\textwidth}
         \centering
         \includegraphics[width=\textwidth]{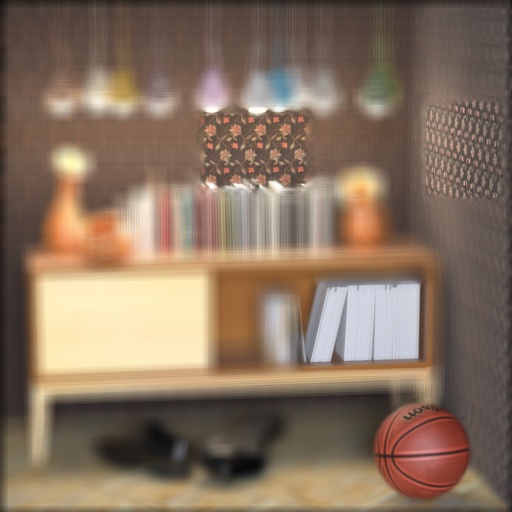}
         \caption{}
     \end{subfigure}
     \hspace{0.1cm}
     \begin{subfigure}[t]{0.29\textwidth}
         \centering
         \includegraphics[width=\textwidth]{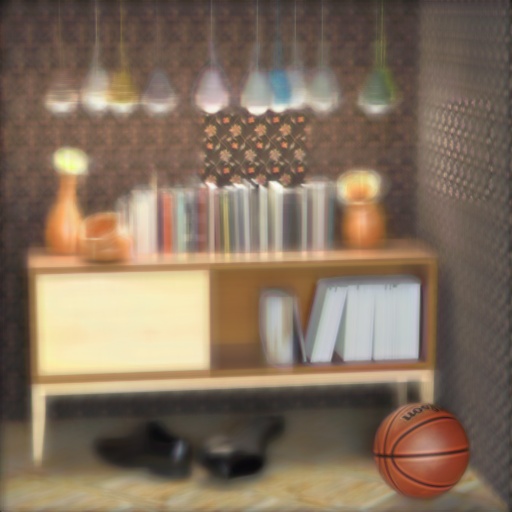}
     \end{subfigure}

    \caption{Multi arbitrary-volume refocusing qualitative results. (a) Lego Knights LF from Stanford LF dataset~\cite{stanfordnewdataset}, (b) Mirabelle Prune Tree, (c) Books LFs from EPFL LF dataset~\cite{epfldataset}, (d) Sideboard LF from HCI LF dataset~\cite{hcidataset}. ROIs in green color boxes $\in \mathfrak{R}_{s}$ and ROIs in red color boxes $\in \mathfrak{R}_{w}$.}
    \label{fig:qualitative}
\end{figure*}

Table \ref{tab:dense_sparse_resultls} shows the computational time and the refocused images' quality measured using BIRSQUE~\cite{brisque} score of the proposed LF refocusing algorithms.
Here, we use $N^{d}_{\alpha}$, $T^{d}_{mask}$, $T^{d}_{refocus}$, $T^{d}_{total}$ to denote the  $\alpha$ value count, $\mathcal{M}_{\alpha}(n_x,n_y)$ generation time, refocused image generation time and total computational time for dense LF refocusing algorithm, respectively. $N^{s}_{\alpha}$, $T^{s}_{mask}$, $T^{s}_{refocus}$, $T^{s}_{total}$ are the counterparts for those in sparse LF refocusing algorithm while $T^{s}_{artifact}$ denotes the time for aliasing artifact removal in sparse LF refocusing algorithm.
$T^{d}_{total}$ and $T^{s}_{total}$ depends on the size (SAI count and resolution) of the LF, $N$, ROI sizes and types ($\in \mathfrak{R}_{w}$ or $\in \mathfrak{R}_{s}$). Lego Knights LF from stanford dataset~\cite{stanfordnewdataset} has has the highest number of SAIs with highest resolution. Therefore, it requires the highest computational time among all tested LFs.  The number of individual refocusings are equal to $N^{d}_{\alpha}$ or $N^{s}_{\alpha}$, and that is quite high due to applying Gaussian filter on $\mathcal{M}_{\alpha}(n_x,n_y)$ for smoothing the transition between regions with different $\alpha$ values. If smooth transition between different regions in $\mathcal{I}_f$ is not applied, $N^{d}_{\alpha}$ and $N^{s}_{\alpha}$ will be limited only to $N$. Therefore, by varying the smoothness according to the user preference, computational time can be reduced. Furthermore, the proposed algorithms utilize the GPU support, significantly reducing the computational time for a single refocusing. As an example, for Lego Knights dense LF, 81 different refocusing requires only 11 seconds limiting single refocusing to less than 0.2 seconds.  For dense LF refocusing algorithm, $T^{d}_{mask}$ is more than $30\%$ of $T^{d}_{total}$. However, $D(n_x,n_y)$ generation is a one time process. Therefore, in applications with dense LFs, $D(n_x,n_y)$ can be generated during setup time and during the processing time, $\mathcal{M}_{\alpha}(n_x,n_y)$ can be generated quickly, reducing the $T^{d}_{mask}$ significantly. The qualitative results are shown in the \figurename~\ref{fig:qualitative}. While dense LF refocusing algorithm perform quite well, the $\mathcal{I}_f$ from sparse LF refocusing algorithm shows some aliasing artifact in Mirabelle Prune Tree LF (\figurename~\ref{fig:mirabelle_prune_tree}). Also there is a slight difference in color and brightness in $\mathcal{I}_f$ of sparse LF refocusing algorithm due to the less number of SAIs and the image restoration CNN. We believe that, these effects can be minimized by training the image restoration CNN further with larger datasets and better augmentations.

\begin{table}[!t]
    \centering
    \caption{Measure refocused image's quality and the computational time of sparse LF refocusing algorithm with respect to that of dense LF refocusing algorithm.}
    \begin{tabular}{l|c >{\centering\arraybackslash}m{1.5cm} l l}
        \hlineB{3}
        \noalign{\vskip 1.5pt}
        \textbf{Light Field}  &  \textbf{SSIM~\cite{ssim}}    &  \textbf{PSNR} & \textbf{Duration}\\ 
        \noalign{\vskip 1.5pt}
        \hlineB{3}
        \noalign{\vskip 1.5pt}
        Lego Knights          & 0.7738  &  24.4413 & 58.7\%\\ 
        Mirablelle Prune Tree & 0.9071  &  24.5378 & 78.2\%\\ 
        Books                 & 0.9342  &  27.9272 & 89.7\%\\ 
        Sideboard             & 0.9230  &  27.8050 & 19.5\%\\ 
        \hlineB{3}
    \end{tabular}
    \label{tab:denseVsSparseCompare}
\end{table}

To evaluate the sparse LF refocusing algorithm, we compared the refocused image quality and the computational time of sparse LF refocusing algorithm with respect to those of dense LF refocusing algorithms, as shown in Table \ref{tab:denseVsSparseCompare}. SSIM value for Lego Knights LF from Standford dataset~\cite{stanfordnewdataset} (captured using a camera array) is relatively low as its SAIs have a larger baseline than LFs from lenselet-based camera, whereas other LFs have SSIM~\cite{ssim} values higher than 0.9. Note that, this much of quality is achieved in sparse LF refocusing algorithm by using less than $20\%$ of SAIs of dense LFs. Furthermore, sparse LF refocusing algorithm takes less computational time than dense LF refocusing algorithm. In the presented LFs and ROIs here, $T^{s}_{total}$ always less than $90\%$ of $T^{d}_{total}$. This is mainly due to the fact that the $\alpha$ search for sparse LFs is much faster than $D(n_x,n_y)$ generation for dense LFs during $\mathcal{M}_{\alpha}(n_x,n_y)$ generation. 

\section{Conclusion and Future Work}\label{sec:conclusion}

We demonstrate interactive post-capture refocusing of dense and sparse LFs with arbitrarily selected single or multiple volumetric regions. Our method employ the shift-and-sum algorithm with pixel-dependent depths to achieve simultaneous in-focus and out-of-focus regions having the same depth range. We almost completely eliminate ghosting artifacts resulted due to the undersampling of sparse LFs by using a U-Net based deep learning model. Experimental results confirm that the proposed method achieves impressive results for both dense and sparse LFs. Furthermore, our method is fast and achieves near real-time processing for sparse LFs. Future work includes design of hardware architectures to efficiently implement the proposed method for real-time interactive LF refocusing. 


\balance
\bibliographystyle{IEEEtran}
\bibliography{references,References_5DLFV}

\begin{thebibliography}{10}
\providecommand{\url}[1]{#1}
\csname url@samestyle\endcsname
\providecommand{\newblock}{\relax}
\providecommand{\bibinfo}[2]{#2}
\providecommand{\BIBentrySTDinterwordspacing}{\spaceskip=0pt\relax}
\providecommand{\BIBentryALTinterwordstretchfactor}{4}
\providecommand{\BIBentryALTinterwordspacing}{\spaceskip=\fontdimen2\font plus
\BIBentryALTinterwordstretchfactor\fontdimen3\font minus
  \fontdimen4\font\relax}
\providecommand{\BIBforeignlanguage}[2]{{%
\expandafter\ifx\csname l@#1\endcsname\relax
\typeout{** WARNING: IEEEtran.bst: No hyphenation pattern has been}%
\typeout{** loaded for the language `#1'. Using the pattern for}%
\typeout{** the default language instead.}%
\else
\language=\csname l@#1\endcsname
\fi
#2}}
\providecommand{\BIBdecl}{\relax}
\BIBdecl

\bibitem{Ade1991}
E.~H. Adelson and J.~R. Bergen, ``The plenoptic function and the elements of
  early vision,'' in \emph{Computation Models of Visual Processing}, M.~Landy
  and J.~A. Movshon, Eds.\hskip 1em plus 0.5em minus 0.4em\relax Cambridge, MA:
  MIT Press, 1991, pp. 3--20.

\bibitem{Zhan2003}
C.~Zhang and T.~Chen, ``Spectral analysis for sampling image-based rendering
  data,'' vol.~13, no.~11, pp. 1038--1050, Nov. 2003.

\bibitem{wanner2013variational}
S.~Wanner and B.~Goldluecke, ``Variational light field analysis for disparity
  estimation and super-resolution,'' \emph{IEEE transactions on pattern
  analysis and machine intelligence}, vol.~36, no.~3, pp. 606--619, 2013.

\bibitem{tao2013depth}
M.~W. Tao, S.~Hadap, J.~Malik, and R.~Ramamoorthi, ``Depth from combining
  defocus and correspondence using light-field cameras,'' in \emph{Proceedings
  of the IEEE International Conference on Computer Vision}, 2013, pp. 673--680.

\bibitem{wang2015depth}
T.-C. Wang, A.~A. Efros, and R.~Ramamoorthi, ``Occlusion-aware depth estimation
  using light-field cameras,'' in \emph{Proceedings of the IEEE International
  Conference on Computer Vision}, 2015, pp. 3487--3495.

\bibitem{vaish2004occlusion}
V.~Vaish, B.~Wilburn, N.~Joshi, and M.~Levoy, ``Using plane+ parallax for
  calibrating dense camera arrays,'' in \emph{Proceedings of the 2004 IEEE
  Computer Society Conference on Computer Vision and Pattern Recognition, 2004.
  CVPR 2004.}, vol.~1.\hskip 1em plus 0.5em minus 0.4em\relax IEEE, 2004, pp.
  I--I.

\bibitem{Dan2007}
D.~Dansereau and L.~T. Bruton, ``A {4-D} dual-fan filter bank for depth
  filtering in light fields,'' vol.~55, no.~2, pp. 542--549, Feb. 2007.

\bibitem{Liyanageocclusion}
N.~Liyanage, C.~Wijenayake, C.~Edussooriya, A.~Madanayake, P.~Agathoklis, L.~T.
  Bruton, and E.~Ambikairajah, ``Multi-depth filtering and occlusion
  suppression in 4-d light fields: Algorithms and architectures,'' \emph{Signal
  Processing}, vol. 167, p. 107294, 2020.

\bibitem{Dong2013}
F.~Dong, S.-H. Ieng, X.~Savatier, R.~Etienne-Cummings, and R.~Benosman,
  ``Plenoptic cameras in real-time robotics,'' \emph{Int. J. Robot. Res.},
  vol.~32, no.~2, pp. 206--217, Feb. 2013.

\bibitem{Tsai2017}
D.~Tsai, D.~G. Dansereau, T.~Peynot, and P.~Corke, ``Image-based visual
  servoing with light field cameras,'' \emph{IEEE Robot. and Automation Lett.},
  vol.~2, no.~2, pp. 912--919, Apr. 2017.

\bibitem{Baj2018}
A.~Bajpayee, A.~H. Techet, and H.~Singh, ``Real-time light field processing for
  autonomous robotics,'' in \emph{Proc. IEEE/RSJ Int. Conf. Intell. Robot.
  Syst.}, 2018, pp. 4218--4225.

\bibitem{Piza2013}
O.~Pizarro, S.~B. Williams, M.~V. Jakuba, M.~Johnson-Roberson, I.~Mahon,
  M.~Bryson, D.~Steinberg, A.~Friedman, D.~Dansereau, N.~Nourani-Vatani
  \emph{et~al.}, ``Benthic monitoring with robotic platforms—the experience
  of australia,'' in \emph{Proc. IEEE Int. Symp. Underwater Technol.}, 2013,
  pp. 1--10.

\bibitem{Dan2014}
D.~G. Dansereau, ``Plenoptic signal processing for robust vision in field
  robotics,'' Ph.D. dissertation, Australian Centre for Field Robotics, School
  of Aerospace, Mechanical and Mechatronic Engineering, 2014.

\bibitem{ng2005light}
R.~Ng, M.~Levoy, M.~Br{\'e}dif, G.~Duval, M.~Horowitz, and P.~Hanrahan, ``Light
  field photography with a hand-held plenoptic camera,'' Ph.D. dissertation,
  Stanford University, 2005.

\bibitem{multivolume}
S.~S. Jayaweera, C.~U. Edussooriya, C.~Wijenayake, P.~Agathoklis, and L.~T.
  Bruton, ``Multi-volumetric refocusing of light fields,'' \emph{IEEE Signal
  Processing Letters}, vol.~28, pp. 31--35, 2020.

\bibitem{volumetric}
D.~G. Dansereau, O.~Pizarro, and S.~B. Williams, ``Linear volumetric focus for
  light field cameras.'' \emph{ACM Trans. Graph.}, vol.~34, no.~2, pp. 15--1,
  2015.

\bibitem{Tro2019}
J.~Trottnow \emph{et~al.}, ``The potential of light fields in media
  productions,'' in \emph{Proc. SIGGRAPH Asia Technical Briefs}, 2019, pp.
  71--74.

\bibitem{LFVideoCamera}
B.~S. Wilburn, M.~Smulski, H.-H.~K. Lee, and M.~A. Horowitz, ``Light field
  video camera,'' in \emph{Media Processors 2002}, vol. 4674.\hskip 1em plus
  0.5em minus 0.4em\relax SPIE, 2001, pp. 29--36.

\bibitem{jiang2018sparse_depth}
X.~Jiang, M.~Le~Pendu, and C.~Guillemot, ``Depth estimation with occlusion
  handling from a sparse set of light field views,'' in \emph{2018 25th IEEE
  International Conference on Image Processing (ICIP)}.\hskip 1em plus 0.5em
  minus 0.4em\relax IEEE, 2018, pp. 634--638.

\bibitem{learning_view_synthesis}
N.~K. Kalantari, T.-C. Wang, and R.~Ramamoorthi, ``Learning-based view
  synthesis for light field cameras,'' \emph{ACM Transactions on Graphics
  (TOG)}, vol.~35, no.~6, pp. 1--10, 2016.

\bibitem{2015fast_real}
C.-T. Huang, J.~Chin, H.-H. Chen, Y.-W. Wang, and L.-G. Chen, ``Fast realistic
  refocusing for sparse light fields,'' in \emph{2015 IEEE International
  Conference on Acoustics, Speech and Signal Processing (ICASSP)}.\hskip 1em
  plus 0.5em minus 0.4em\relax IEEE, 2015, pp. 1176--1180.

\bibitem{alain2021spatio_sparse}
M.~Alain and A.~Smolic, ``A spatio-angular filter for high quality sparse light
  field refocusing,'' in \emph{2021 IEEE International Conference on Multimedia
  \& Expo Workshops (ICMEW)}.\hskip 1em plus 0.5em minus 0.4em\relax IEEE,
  2021, pp. 1--6.

\bibitem{unet}
O.~Ronneberger, P.~Fischer, and T.~Brox, ``U-net: Convolutional networks for
  biomedical image segmentation,'' in \emph{International Conference on Medical
  image computing and computer-assisted intervention}.\hskip 1em plus 0.5em
  minus 0.4em\relax Springer, 2015, pp. 234--241.

\bibitem{ng2005fourier}
R.~Ng, ``Fourier slice photography,'' in \emph{ACM Siggraph 2005 Papers}, 2005,
  pp. 735--744.

\bibitem{tilt_shift}
M.~Alain, W.~Aenchbacher, and A.~Smolic, ``Interactive light field tilt-shift
  refocus with generalized shift-and-sum,'' \emph{arXiv preprint
  arXiv:1910.04699}, 2019.

\bibitem{volumeSparseFIR}
S.~U. Premaratne, C.~U. Edussooriya, C.~Wijenayake, L.~T. Bruton, and
  P.~Agathoklis, ``A 4-d sparse fir hyperfan filter for volumetric refocusing
  of light fields by hard thresholding,'' in \emph{2018 IEEE 23rd International
  Conference on Digital Signal Processing (DSP)}.\hskip 1em plus 0.5em minus
  0.4em\relax IEEE, 2018, pp. 1--5.

\bibitem{2021unsupervised}
S.~T. Digumarti, J.~Daniel, A.~Ravendran, R.~Griffiths, and D.~G. Dansereau,
  ``Unsupervised learning of depth estimation and visual odometry for sparse
  light field cameras,'' in \emph{2021 IEEE/RSJ International Conference on
  Intelligent Robots and Systems (IROS)}.\hskip 1em plus 0.5em minus
  0.4em\relax IEEE, 2021, pp. 278--285.

\bibitem{deep_sparse}
S.~B. Dayan, D.~Mendlovic, and R.~Giryes, ``Deep sparse light field
  refocusing,'' \emph{arXiv preprint arXiv:2009.02582}, 2020.

\bibitem{machinelearning_multi_refocus}
E.~Hedayati, T.~C. Havens, and J.~P. Bos, ``Machine learning method for light
  field refocusing,'' \emph{arXiv preprint arXiv:2103.16020}, 2021.

\bibitem{levoy1996light}
M.~Levoy and P.~Hanrahan, ``Light field rendering,'' in \emph{Proceedings of
  the 23rd annual conference on Computer graphics and interactive techniques},
  1996, pp. 31--42.

\bibitem{ms_ssim}
Z.~Wang, E.~P. Simoncelli, and A.~C. Bovik, ``Multiscale structural similarity
  for image quality assessment,'' in \emph{The Thrity-Seventh Asilomar
  Conference on Signals, Systems \& Computers, 2003}, vol.~2.\hskip 1em plus
  0.5em minus 0.4em\relax Ieee, 2003, pp. 1398--1402.

\bibitem{epfldataset}
M.~Rerabek and T.~Ebrahimi, ``New light field image dataset,'' in \emph{8th
  International Conference on Quality of Multimedia Experience (QoMEX)}, no.
  CONF, 2016.

\bibitem{hinet}
L.~Chen, X.~Lu, J.~Zhang, X.~Chu, and C.~Chen, ``Hinet: Half instance
  normalization network for image restoration,'' in \emph{Proceedings of the
  IEEE/CVF Conference on Computer Vision and Pattern Recognition}, 2021, pp.
  182--192.

\bibitem{mprnet}
S.~W. Zamir, A.~Arora, S.~Khan, M.~Hayat, F.~S. Khan, M.-H. Yang, and L.~Shao,
  ``Multi-stage progressive image restoration,'' in \emph{Proceedings of the
  IEEE/CVF Conference on Computer Vision and Pattern Recognition}, 2021, pp.
  14\,821--14\,831.

\bibitem{swinir}
J.~Liang, J.~Cao, G.~Sun, K.~Zhang, L.~Van~Gool, and R.~Timofte, ``Swinir:
  Image restoration using swin transformer,'' in \emph{Proceedings of the
  IEEE/CVF International Conference on Computer Vision}, 2021, pp. 1833--1844.

\bibitem{psnr}
A.~Hore and D.~Ziou, ``Image quality metrics: Psnr vs. ssim,'' in \emph{2010
  20th international conference on pattern recognition}.\hskip 1em plus 0.5em
  minus 0.4em\relax IEEE, 2010, pp. 2366--2369.

\bibitem{stanfordnewdataset}
V.~Vaish and A.~Adams, ``The (new) stanford light field archive,''
  \emph{Computer Graphics Laboratory, Stanford University}, vol.~6, no.~7,
  p.~3, 2008.

\bibitem{hcidataset}
K.~Honauer, O.~Johannsen, D.~Kondermann, and B.~Goldluecke, ``A dataset and
  evaluation methodology for depth estimation on 4d light fields,'' in
  \emph{Asian Conference on Computer Vision}.\hskip 1em plus 0.5em minus
  0.4em\relax Springer, 2016, pp. 19--34.

\bibitem{xavier}
X.~Glorot and Y.~Bengio, ``Understanding the difficulty of training deep
  feedforward neural networks,'' in \emph{Proceedings of the thirteenth
  international conference on artificial intelligence and statistics}.\hskip
  1em plus 0.5em minus 0.4em\relax JMLR Workshop and Conference Proceedings,
  2010, pp. 249--256.

\bibitem{brisque}
A.~Mittal, A.~K. Moorthy, and A.~C. Bovik, ``No-reference image quality
  assessment in the spatial domain,'' \emph{IEEE Transactions on Image
  Processing}, vol.~21, no.~12, pp. 4695--4708, 2012.

\bibitem{ssim}
Z.~Wang, A.~C. Bovik, H.~R. Sheikh, and E.~P. Simoncelli, ``Image quality
  assessment: from error visibility to structural similarity,'' \emph{IEEE
  transactions on image processing}, vol.~13, no.~4, pp. 600--612, 2004.

\end{thebibliography}


\end{document}